\documentclass[letterpaper]{article}
\usepackage[preprint]{aaai2027}
\usepackage[hyphens]{url}
\usepackage{graphicx}
\usepackage{natbib}
\usepackage{caption}
\usepackage{amsmath}
\usepackage{amssymb}
\usepackage{booktabs}
\usepackage{multirow}

\graphicspath{{figures/}}
\title{Spatial Information Bottleneck for Interpretable Image Classification}
\author{Kaixiang Shu}
\affiliations{Independent Researcher}

\begin{document}
\maketitle

\begin{abstract}
Post-hoc explainers can reveal only evidence encoded by a classifier, yet cross-entropy leaves the spatial distribution of predictive sensitivity unconstrained. We study the vector--Jacobian product (VJP) of posterior concentration as a training-time representation. Within a fixed piecewise-affine region, a spatially restricted VJP locally determines the model posterior under explicit tangent-injectivity and nonsingularity conditions. If the model posterior equals the Bayes posterior, this VJP is minimal sufficient for the corresponding conditional prediction experiment. Within that neighborhood, expected posterior KL upper-bounds the remaining label-information gap. This analysis motivates Spatial Information Bottleneck (S-IB), which requires no spatial or concept annotations beyond class labels. S-IB derives a soft VJP mask, aligns foreground input--VJP similarity structures using centered linear-kernel HSIC, and suppresses coordinate-centered background VJP variation. Across five datasets and four backbones, the matched S-IB point estimate is higher in all 20 top-1 comparisons by 0.30--2.52 points. On ResNet-50, all six explainers have positive mean localization changes; mean Insertion is also higher in all 12 fine-grained backbone--dataset settings. Deletion is dataset- and architecture-dependent and worsens for all seven signals on Oxford-IIIT Pet/ResNet-50. Inference is unchanged, although second-order training requires about $2.4\times$ time and $2.0\times$ memory. Overall, S-IB improves several reported outcomes for unchanged explainers within explicit theoretical, metric, and cost boundaries.
 \end{abstract}

\section{Introduction}
\label{sec:introduction}

\begin{figure*}[t]
\centering
\includegraphics[width=0.94\textwidth]{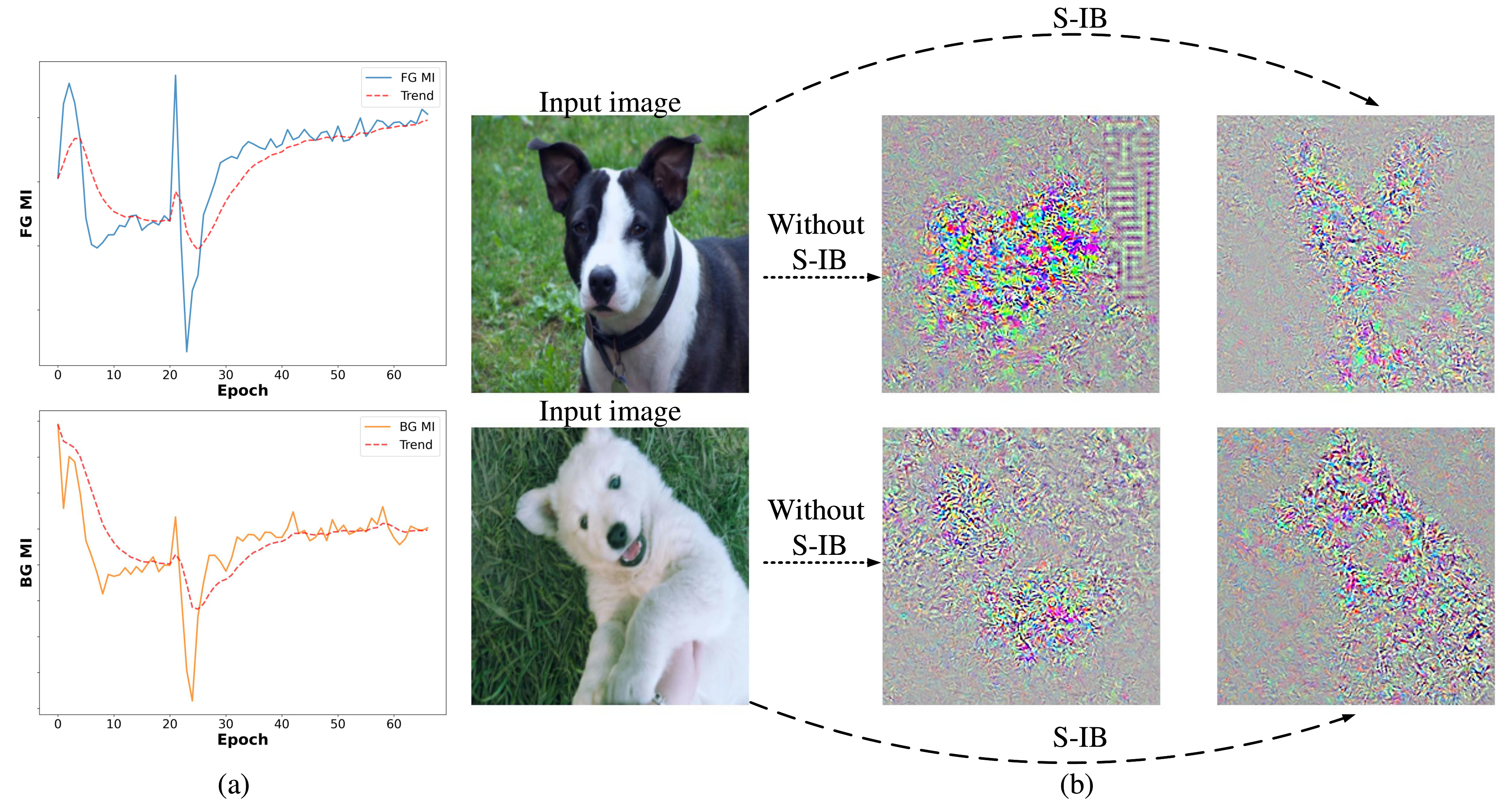}
\caption{S-IB motivation. (a) Foreground/background training traces; the embedded ``MI'' label denotes HSIC dependence, not mutual information. (b) Illustrative attributions before and after S-IB; Tables~\ref{tab:localization_delta}--\ref{tab:insertion} report localization and Insertion results.}
\label{fig:original_motivation}
\end{figure*}

Post-hoc attribution is widely used to inspect image classifiers, but an explainer can expose only spatial evidence that the classifier encoded. Gradient- and activation-based methods can therefore remain diffuse or background-sensitive even when classification accuracy is high~\cite{simonyan2013deep,selvaraju2017grad,sundararajan2017axiomatic,wang2020score,colin2022cannot,weber2023beyond}. This gap matters because a plausible heatmap need not faithfully identify the evidence used for prediction~\cite{arrieta2020explainable,janzing2020feature}. Improving the explainer alone leaves that evidence representation unchanged. Figure~\ref{fig:original_motivation} summarizes this empirical gap.

Training-time approaches reshape representations but address adjacent settings. Concept bottlenecks and prototypes often require semantic labels, architecture changes, or both~\cite{koh2020concept,yuksekgonul2022post,chen2019looks,rymarczyk2022interpretable}. Right-for-the-Right-Reasons constrains sensitivity using annotated irrelevant regions~\cite{ross2017right}, while shortcut methods commonly use group annotations or multiple environments~\cite{liu2021just,kirichenko2022last,zhang2022correct}. Information Bottleneck (IB) objectives regularize representation complexity~\cite{tishby2000information,alemi2016deep,fischer2020conditional}, but not the location of posterior sensitivity. We instead shape spatial evidence using only class labels, without changing the classifier, attribution rule, or inference path.

We address this gap by asking when a spatially restricted VJP preserves the predictive information in the model posterior. Within a fixed piecewise-affine region, the restricted VJP locally recovers that posterior under explicit Jacobian-injectivity and nonsingularity conditions. This equivalence becomes statistical minimal sufficiency when the model posterior equals the Bayes posterior. Expected posterior KL controls the information gap otherwise. This local retention criterion motivates Spatial Information Bottleneck (S-IB), which derives a deterministic soft mask from the VJP. S-IB aligns cross-sample similarity structures of masked inputs and VJPs while suppressing coordinate-centered variation in the complementary branch. The paired foreground--background decomposition remains exactly reconstructible.

We evaluate S-IB on five datasets, four backbones, and six post-hoc explainers. Its matched top-1 point estimate is higher in all 20 dataset--architecture comparisons. On ResNet-50 across three fine-grained datasets, every explainer has a positive mean localization change; mean Insertion is higher for all 12 tested backbone--dataset pairs. Component controls favor both regularizers and VJP-based mask targeting in the tested setting. The evaluation also exposes clear boundaries. Evidence is limited by class-supervised evaluation, heterogeneous Deletion, and second-order training cost.

Our contributions are threefold:
\begin{itemize}
    \item We establish local equivalence between a fixed spatial restriction of the VJP and the model posterior, then derive minimal sufficiency when that posterior is correct. We separately define inclusion-minimal spatial support and bound the approximate information gap by posterior KL.
    \item We introduce a spatial information bottleneck requiring no spatial or concept annotations beyond class labels. It combines a deterministic VJP mask with cross-sample foreground dependence alignment and background coordinate-variation suppression, without modifying the classifier or inference path.
    \item We evaluate five datasets, four backbones, and six post-hoc explainers, together with component, sensitivity, theoretical-condition, and cost analyses.
\end{itemize}
 \section{Related Work}

\subsection{Post-Hoc Attribution}
Gradient explanations estimate input sensitivity or integrate it along paths~\cite{simonyan2013deep,smilkov2017smoothgrad,sundararajan2017axiomatic,springenberg2014striving}. CAM variants combine feature maps with class-dependent weights~\cite{zhou2016learning,selvaraju2017grad,chattopadhay2018grad}; ScoreCAM uses forward activation scores~\cite{wang2020score}. Such maps may be fragile or weakly coupled to model behavior~\cite{weber2023beyond,hase2023does}. S-IB changes training instead, and we evaluate unchanged attribution rules on the resulting models.

\subsection{Training-Time Interpretability}
Concept bottlenecks expose human-defined concepts~\cite{koh2020concept,yuksekgonul2022post,oikarinen2023label}, while prototype networks use learned exemplars~\cite{chen2019looks,rymarczyk2022interpretable}. A close objective-level precedent is Right-for-the-Right-Reasons: it also constrains input derivatives but requires externally specified irrelevant features~\cite{ross2017right}. S-IB derives its soft split from the model's posterior-concentration VJP, uses only class labels, and preserves the original classifier. Shortcut-mitigation methods based on sampling, reweighting, or classifier fitting address a complementary supervision setting~\cite{liu2021just,kirichenko2022last,zhang2022correct}.

\subsection{Information-Theoretic Representation Learning}
The IB principle balances predictive information against representation complexity~\cite{tishby2000information}; variational and conditional formulations make it tractable for neural networks~\cite{alemi2016deep,fischer2020conditional}. Their standard objectives act on a latent representation as a whole. S-IB splits an input-space posterior VJP into reconstructible spatial branches and applies asymmetric surrogates. Foreground alignment uses centered linear-kernel HSIC~\cite{gretton2005measuring,song2007supervised,gretton2007kernel}; the background term controls a Gaussian-smoothed dependence bound through coordinate-centered variation. Thus, ``information bottleneck'' names a spatial retain--suppress design rather than an estimator of classical IB quantities.
 \section{Spatial Information Bottleneck}
\label{sec:method}

\subsection{Overview and VJP Representation}
Let $x\in\mathbb{R}^{c\times h\times w}$ be an image and $y\in\{1,\ldots,C\}$ its label. A classifier produces logits $z_\theta(x)$ and model posterior $p_\theta(x)=\operatorname{softmax}(z_\theta(x)/\tau)$, where $\tau>0$. All experiments use the standard choice $\tau=1$. Cross-entropy determines which evidence predicts the label but not where its sensitivity lies. We define the posterior-concentration VJP
\begin{equation}
R_\theta(x)=J_p(x)^\top p_\theta(x)
=\nabla_x \frac{1}{2}\lVert p_\theta(x)\rVert_2^2,
\label{eq:vjp}
\end{equation}
where $J_p(x)=\partial p_\theta(x)/\partial x$. Equation~\ref{eq:vjp} matches the implementation and produces a tensor with the same shape as $x$. It measures how local input perturbations change posterior concentration. This training representation is distinct from the class-specific targets used by post-hoc explainers. S-IB regularizes Eq.~\ref{eq:vjp}, while evaluation applies each explainer with its standard target.

Equation~\ref{eq:vjp} is target-free: it weights class-wise posterior derivatives by the complete model posterior rather than selecting a ground-truth or predicted class. One reverse pass produces an input-shaped tensor, so the same representation can drive spatial regularization for every training sample. These properties make the VJP a differentiable carrier of classifier evidence, not an explanation rule by itself.

\begin{figure}[t]
\centering
\includegraphics[width=\columnwidth]{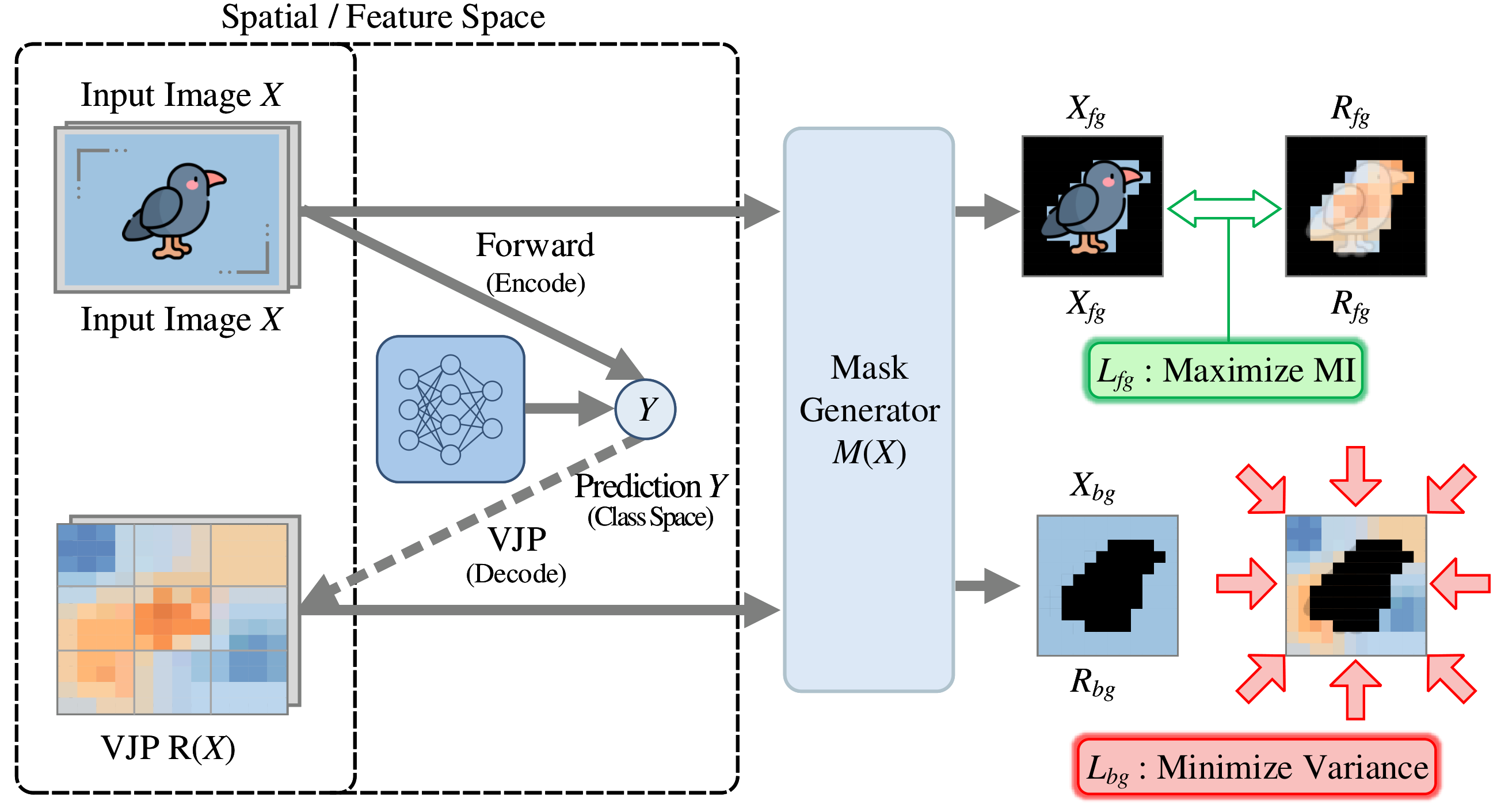}
\caption{S-IB workflow. The embedded $M(X)$ denotes deterministic $M=\phi(R)$, while ``maximize MI'' denotes centered linear-kernel HSIC, not mutual-information estimation. The regularization branch is discarded at inference.}
\label{fig:framework}
\end{figure}

Figure~\ref{fig:framework} summarizes one S-IB update. A forward pass produces $p_\theta(x)$ and the classification loss, and one reverse-mode VJP produces $R_\theta(x)$. The deterministic operator $\phi$ constructs a soft spatial mask. Foreground and background regularizers then update the original classifier through the retained VJP graph. No explanation module is trained, and the S-IB regularization branch is removed after optimization.

\subsection{When Is the VJP Minimally Sufficient?}
A statistic $T=T(X)$ is sufficient for predicting the label when $Y\perp X\mid T$. The Bayes posterior $\eta(x)=P(Y=\cdot\mid X=x)$ is sufficient, and any sufficient $T$ obeys $\eta(X)=P(Y=\cdot\mid T)$ almost surely. Thus, $\eta$ is a function of every sufficient statistic and is minimal up to almost-sure equivalence~\cite{blackwell1953equivalent}. We establish when the VJP is locally equivalent to the model posterior, and hence to $\eta$ when that posterior is correct.

Consider a fixed activation region $\mathcal{U}$ of a piecewise-affine network, so $z_\theta(x)=Ax+b$ for $x\in\mathcal{U}$. Let $\mathcal{M}$ be a set of spatial positions, expanded across all input channels, and let $A_{\mathcal{M}}$ be the logit Jacobian restricted to those coordinates. The restricted VJP factors as
\begin{equation}
g(p)=\frac{1}{\tau}\left(\operatorname{diag}(p)-pp^\top\right)p,
\quad
R_{\mathcal{M}}=A_{\mathcal{M}}^\top g(p).
\label{eq:vjp_factor}
\end{equation}
Indeed, $J_p=\tau^{-1}(\operatorname{diag}(p)-pp^\top)A$, so Eq.~\ref{eq:vjp_factor} separates posterior geometry, represented by $g(p)$, from the fixed local logit geometry $A_{\mathcal M}$. The vector $g(p)$ lies in the simplex tangent space $\mathcal{T}=\{v:\mathbf{1}^\top v=0\}$. Its differential on $\mathcal{T}$ is
\begin{equation}
Dg_p[v]=\frac{1}{\tau}B_pv,
\quad
B_p=2\operatorname{diag}(p)-2pp^\top-\lVert p\rVert_2^2I.
\label{eq:vjp_differential}
\end{equation}

\noindent\textbf{Theorem 1 (Local Posterior Equivalence and Conditional Minimal Sufficiency).}
Fix $x_0\in\mathcal{U}$ with $p_0=p_\theta(x_0)$ in the simplex interior. Assume (i) $A_{\mathcal{M}}^\top$ is injective on $\mathcal{T}$, equivalently $\ker(A_{\mathcal{M}}^\top)\cap\mathcal{T}=\{0\}$; and (ii) $B_{p_0}$ is nonsingular on $\mathcal{T}$. Then a neighborhood $\mathcal{V}\subseteq\mathcal{U}$ of $x_0$ and a smooth map $h$ exist such that $h(R_{\mathcal{M}}(x))=p_\theta(x)$ for every $x\in\mathcal{V}$. If $p_\theta(x)=\eta(x)$ on $\mathcal{V}$, then $R_{\mathcal{M}}(X)$ is minimal sufficient for $Y$ under the law conditioned on $X\in\mathcal{V}$.

\noindent\textit{Proof sketch.}
Within $\mathcal{U}$, $A_{\mathcal{M}}$ is constant, so $R_{\mathcal{M}}=F(p)$ depends on $x$ only through $p_\theta(x)$. Equations~\ref{eq:vjp_factor}--\ref{eq:vjp_differential} give $DF_{p_0}=A_{\mathcal{M}}^\top B_{p_0}/\tau$. Assumptions (i) and (ii) give full rank $C-1$ on $\mathcal{T}$. This rank persists locally, so the constant-rank theorem makes $F$ a local embedding with a smooth left inverse on its image. Because $R_{\mathcal{M}}$ is also a deterministic function of $p$, they generate the same information on $\mathcal{V}$. If $p=\eta$, $R_{\mathcal{M}}$ determines the conditional label law and is sufficient. Any sufficient $T(X)$ under the conditional law determines $\eta(X)=P(Y=\cdot\mid T(X),X\in\mathcal{V})$ almost surely. Thus, every sufficient $T$ determines $R_{\mathcal{M}}$, establishing minimality.

\noindent\textbf{Remark 1 (Genericity is not conditioning).}
At the uniform posterior, $B_pv=C^{-1}v$ on $\mathcal{T}$. Therefore, $\det(B_p|_{\mathcal T})$ is not the zero polynomial, and singular posteriors form a lower-dimensional set. This establishes that condition (ii) is non-vacuous, but it does not guarantee numerical stability: a nearly singular $B_p$ can make the local inverse ill-conditioned even when it exists.

\noindent\textbf{Corollary 1 (Approximate Sufficiency).}
Let $\mathcal{V}$ be a neighborhood on which the left inverse in Theorem~1 exists. Let $I_{\mathcal{V}}$ and $\mathbb{E}_{\mathcal{V}}$ denote information and expectation under the law conditioned on $X\in\mathcal{V}$, and let $q(u)=P(Y=\cdot\mid p_\theta(X)=u,X\in\mathcal{V})$. Then
\begin{equation}
\begin{aligned}
&I_{\mathcal{V}}(X;Y)-I_{\mathcal{V}}(R_{\mathcal{M}};Y)\\
&=\mathbb{E}_{\mathcal{V}}D_{\mathrm{KL}}\!\left(\eta(X)\Vert q(p_\theta(X))\right)\\
&\leq \mathbb{E}_{\mathcal{V}}D_{\mathrm{KL}}\!\left(\eta(X)\Vert p_\theta(X)\right).
\end{aligned}
\label{eq:approx_sufficiency}
\end{equation}
The equality follows because $R_{\mathcal{M}}$ and $p_\theta$ determine each other on $\mathcal{V}$. The conditional posterior $q$ minimizes expected log loss among functions of $p_\theta(X)$; using $p_\theta(X)$ itself gives the inequality. This separates geometric recoverability from posterior approximation error. The right-hand side measures posterior correctness rather than standard calibration, and our experiments do not estimate it.

Spatial minimality is a separate local notion. Over the finite grid $\Omega$, let $\mathfrak{A}(x_0,\mathcal{U})$ contain supports satisfying both geometric conditions. Any nonempty family has an inclusion-minimal element $\mathcal{M}^\star(x_0,\mathcal{U})$ whose proper subsets fail local recoverability; this element may change across regions. The theorem fixes $\mathcal{M}$ throughout $\mathcal{V}$ and gives a retention criterion, not an estimator of $\mathcal{M}^\star$ or the input-dependent $M(x)=\phi(R(x))$. S-IB instead uses a jointly reconstructible soft split and empirically audits foreground-restricted Jacobian rank. The exact inverse concerns a frozen piecewise-affine map; smooth nonlinearities and training-time batch coupling admit only a first-order local interpretation.

This distinction links theory and algorithm. Theorem~1 gives a fixed-support retention criterion; S-IB uses $\phi(R)$ to relax input-dependent support selection. The foreground objective promotes input--VJP agreement on retained coordinates, while the background objective suppresses complementary variation. Equation~\ref{eq:split} preserves exact reconstruction.

\subsection{Deterministic Soft Spatial Mask}
The mask generator $\phi$ in Figure~\ref{fig:framework} has no learned parameters. It begins with the nonnegative spatial response
\begin{equation}
S=\left|\frac{1}{c}\sum_{j=1}^{c}R_{\theta,j}(x)\right|.
\end{equation}
Let $s_{(0)}\leq\cdots\leq s_{(n-1)}$ be the sorted entries of a spatial tensor. The implementation replaces a hard order statistic by
\begin{equation}
\begin{aligned}
Q_q(s)&=\sum_{i=0}^{n-1}w_i(q)s_{(i)},\\
w_i(q)&=\frac{\exp\!\left(-|i-q(n-1)|/\tau_q\right)}
{\sum_{\ell=0}^{n-1}\exp\!\left(-|\ell-q(n-1)|/\tau_q\right)}.
\end{aligned}
\label{eq:soft_quantile}
\end{equation}
with sorting temperature $\tau_q=0.1$. Thus, the weights concentrate around the target rank rather than imposing a dataset-level response threshold. We then apply $\sigma_a(u)=\operatorname{sigmoid}(au)$:
\begin{equation}
\begin{aligned}
\widehat S&=\frac{S-Q_{0.05}(S)}{Q_{0.95}(S)-Q_{0.05}(S)+\epsilon},\\
P&=\sigma_{10}\!\left(\widehat S-Q_{0.8}(\widehat S)\right),\\
D&=\operatorname{AvgPool}_{31\times31}(P),\qquad
M=\sigma_{10}\!\left(D-Q_{0.8}(D)\right).
\end{aligned}
\label{eq:mask}
\end{equation}
The operator is piecewise differentiable away from sorting ties. The 5th--95th percentile normalization reduces sensitivity to samplewise scale and offset. The first gate selects relatively large responses, average pooling aggregates local support, and the second gate reapplies a relative spatial threshold. The resulting mask has no learned parameters and is broadcast across input channels.

\subsection{Spatial Decomposition and IB Surrogates}
The mask defines
\begin{equation}
\begin{aligned}
R_{\mathrm{fg}}&=R\odot M, &R_{\mathrm{bg}}&=R\odot(1-M),\\
X_{\mathrm{fg}}&=X\odot M, &X_{\mathrm{bg}}&=X\odot(1-M).
\end{aligned}
\label{eq:split}
\end{equation}
\noindent\textbf{Exact split identity.}
For every $M\in[0,1]^{h\times w}$, Eq.~\ref{eq:split} satisfies $R_{\mathrm{fg}}+R_{\mathrm{bg}}=R$ and $X_{\mathrm{fg}}+X_{\mathrm{bg}}=X$. Each branch pair therefore reconstructs its original tensor by summation. This identity requires neither a binary mask nor statistical independence; the branch-specific objectives alone impose the retain--suppress bias.

For a batch of size $B$, let $U$ and $V$ contain the flattened, $\ell_2$-normalized foreground inputs and VJPs as rows. We form the linear Gram matrices $K=UU^\top$ and $L=VV^\top$, with $H=I-B^{-1}\mathbf{1}\mathbf{1}^{\top}$. The foreground dependence is
\begin{equation}
D_{\mathrm{fg}}=\frac{1}{(B-1)^2}\operatorname{tr}(KHLH).
\label{eq:hsic}
\end{equation}
This statistic aligns cross-sample geometry: pairs close in foreground-input space are encouraged to remain close in foreground-VJP space. It measures dependence rather than mutual information; its value may also reflect the shared gate.

For the background branch, let $r_i=\operatorname{vec}(R_{\mathrm{bg}}^{(i)})\in\mathbb{R}^{d}$, where $d=chw$, and let $C_d=I-d^{-1}\mathbf{1}\mathbf{1}^{\top}$. We use the coordinate-centered loss
\begin{equation}
V_{\mathrm{bg}}=\frac{1}{B(d-1)}\sum_{i=1}^{B}\lVert C_dr_i\rVert_2^2,
\label{eq:variance}
\end{equation}
the mean per-sample variance over vectorized channel--spatial coordinates. Let $\widetilde R_{\mathrm{bg}}=C_d\operatorname{vec}(R_{\mathrm{bg}})$ and let $\widetilde\Sigma$ be its covariance. Define $\overline V_{\mathrm{bg}}=\mathbb{E}[\lVert C_dr\rVert_2^2/(d-1)]$. Assuming a finite second moment, let $N\sim\mathcal{N}(0,\sigma^2I_d)$ be independent. The Markov chain $X_{\mathrm{bg}}\to\widetilde R_{\mathrm{bg}}\to\widetilde R_{\mathrm{bg}}+N$, followed by the Gaussian maximum-entropy bound, yields
\begin{equation}
\begin{aligned}
I(X_{\mathrm{bg}};\widetilde R_{\mathrm{bg}}+N)
&\leq \frac{1}{2}\log\det\!\left(I+\frac{\widetilde\Sigma}{\sigma^2}\right)\\
&\leq \frac{\operatorname{tr}(\widetilde\Sigma)}{2\sigma^2}\\
&\leq \frac{(d-1)\overline V_{\mathrm{bg}}}{2\sigma^2}.
\end{aligned}
\label{eq:bound}
\end{equation}
The second inequality uses $\log\det(I+A)\leq\operatorname{tr}(A)$ for positive semidefinite $A$. The last follows from $\operatorname{tr}(\widetilde\Sigma)\leq\mathbb{E}\lVert\widetilde R_{\mathrm{bg}}\rVert_2^2=(d-1)\overline V_{\mathrm{bg}}$, because subtracting a mean cannot increase the second moment. Thus, $V_{\mathrm{bg}}$ controls Gaussian-smoothed dependence for the coordinate-centered background VJP, but not its global all-ones component. Theorem~1 gives a fixed-support retention criterion; Eqs.~\ref{eq:hsic}--\ref{eq:bound} provide trainable surrogates for the input-dependent split.

\subsection{Training Objective}
We optimize cross-entropy with positive exponential transforms of both regularizers:
\begin{equation}
\mathcal{L}=\mathcal{L}_{\mathrm{ce}}
+\lambda_{\mathrm{bg}}\exp(V_{\mathrm{bg}})
+\lambda_{\mathrm{fg}}\exp(-D_{\mathrm{fg}}).
\label{eq:objective}
\end{equation}
Both exponentials have the intended gradient signs:
\begin{equation}
\frac{\partial\mathcal L}{\partial V_{\mathrm{bg}}}
=\lambda_{\mathrm{bg}}e^{V_{\mathrm{bg}}}>0,
\qquad
\frac{\partial\mathcal L}{\partial D_{\mathrm{fg}}}
=-\lambda_{\mathrm{fg}}e^{-D_{\mathrm{fg}}}<0.
\label{eq:objective_direction}
\end{equation}
Gradient descent therefore suppresses background variation and increases foreground dependence, while the exponentials adapt their gradient magnitudes. We use a background weight of 1000 and a foreground weight of 0.01.

Each minibatch executes five operations. First, it computes logits, posterior, and cross-entropy. Second, it obtains $R$ with one reverse-mode VJP while retaining its graph. Third, it constructs $M=\phi(R)$ and the four tensors in Eq.~\ref{eq:split}. Fourth, it evaluates $D_{\mathrm{fg}}$, $V_{\mathrm{bg}}$, and Eq.~\ref{eq:objective}. Finally, it backpropagates through the VJP graph to update the original classifier. Baseline and S-IB runs otherwise share data, initialization, optimizer, and schedule. For ViT-B/32, the same input-pixel VJP is obtained through the patch-embedding layer, without a token-specific mask path. The full class-wise input Jacobian is never materialized, but differentiating through the VJP requires second-order autograd and dominates the added training cost. Inference retains the original architecture and attribution rule, adding no operations or parameters.
 \section{Experiments}
\label{sec:experiments}

\subsection{Evaluation Protocol}
We use CIFAR-100~\cite{krizhevsky2009learning}, ImageNet~\cite{deng2009imagenet}, CUB-200-2011~\cite{wah2011caltech}, Oxford-IIIT Pet~\cite{parkhi2012cats}, and Stanford Dogs~\cite{khosla2011novel}. The first two test lower-resolution and large-scale settings; the fine-grained datasets also support localization. Backbones are ResNet-18, ResNet-50~\cite{he2016deep}, DenseNet-121~\cite{huang2017densely}, and ViT-B/32~\cite{dosovitskiy2020image}. Every S-IB/baseline pair shares the split, preprocessing, initialization, optimizer, and schedule. Because some settings lack repeated-run variance, differences are descriptive matched point estimates, not significance claims.

We evaluate Saliency~\cite{simonyan2013deep}, Guided Backpropagation~\cite{springenberg2014striving}, Integrated Gradients~\cite{sundararajan2017axiomatic}, GradCAM~\cite{selvaraju2017grad}, GradCAM++~\cite{chattopadhay2018grad}, and ScoreCAM~\cite{wang2020score}. Primary localization comparisons use these six methods with bounding-box Pixel Accuracy, mIoU, and mAP; the supplement also retains direct-VJP localization results. Faithfulness averages the six methods plus Direct VJP. Insertion and Deletion are AUCs, scaled by 100, for revealing or removing pixels in descending attribution order; higher Insertion and lower Deletion are preferred. The supplement provides proof details, complete Top-1/Top-5 classification results, per-signal localization and faithfulness values across architectures, and additional qualitative and diagnostic analyses.

\subsection{Diagnostics for the Theoretical Conditions}
Theorem~1 requires tangent-space injectivity and a nonsingular posterior-to-VJP differential. We probe only injectivity using foreground-restricted ResNet-50 Jacobians from 500 images per dataset. Full row rank $C$ is a conservative sufficient test; condition (ii) is not evaluated.

\begin{table*}[t]
\centering
{\small
\begin{tabular}{lrrrrr}
\toprule
Dataset & Classes & Mask pixels & Effective rank & Rank/$C$ & Rank $=C$ \\
\midrule
CUB-200 & 200 & $8171\pm5154$ & $192.4\pm14.7$ & 96.2\% & 80.2\% (401/500) \\
Oxford-IIIT Pet & 37 & $43816\pm2220$ & $37.0\pm0.0$ & 100.0\% & 100.0\% (500/500) \\
Stanford Dogs & 120 & $28283\pm11716$ & $119.6\pm2.7$ & 99.7\% & 95.4\% (477/500) \\
\bottomrule
\end{tabular}
}
\caption{Foreground-restricted Jacobian diagnostics over 500 images. ``Rank $=C$'' is stricter than tangent-space injectivity.}
\label{tab:rank}
\end{table*}

Effective rank reaches at least 96.2\% of the class dimension, and full row rank holds for 80.2--100\% of images. Because numerical rank is threshold-dependent and rank $C-1$ omits null-space orientation, this supports but does not establish injectivity; the other theorem conditions remain untested.

Existing HSIC traces further audit the split. Within-region estimates lie near $0.1$--$0.4$, while both cross-region terms remain below $0.01$, over an order of magnitude smaller. This supports approximate separation, not independence, and does not exclude shared-gate dependence.

\begin{figure}[t]
\centering
\includegraphics[width=0.98\columnwidth]{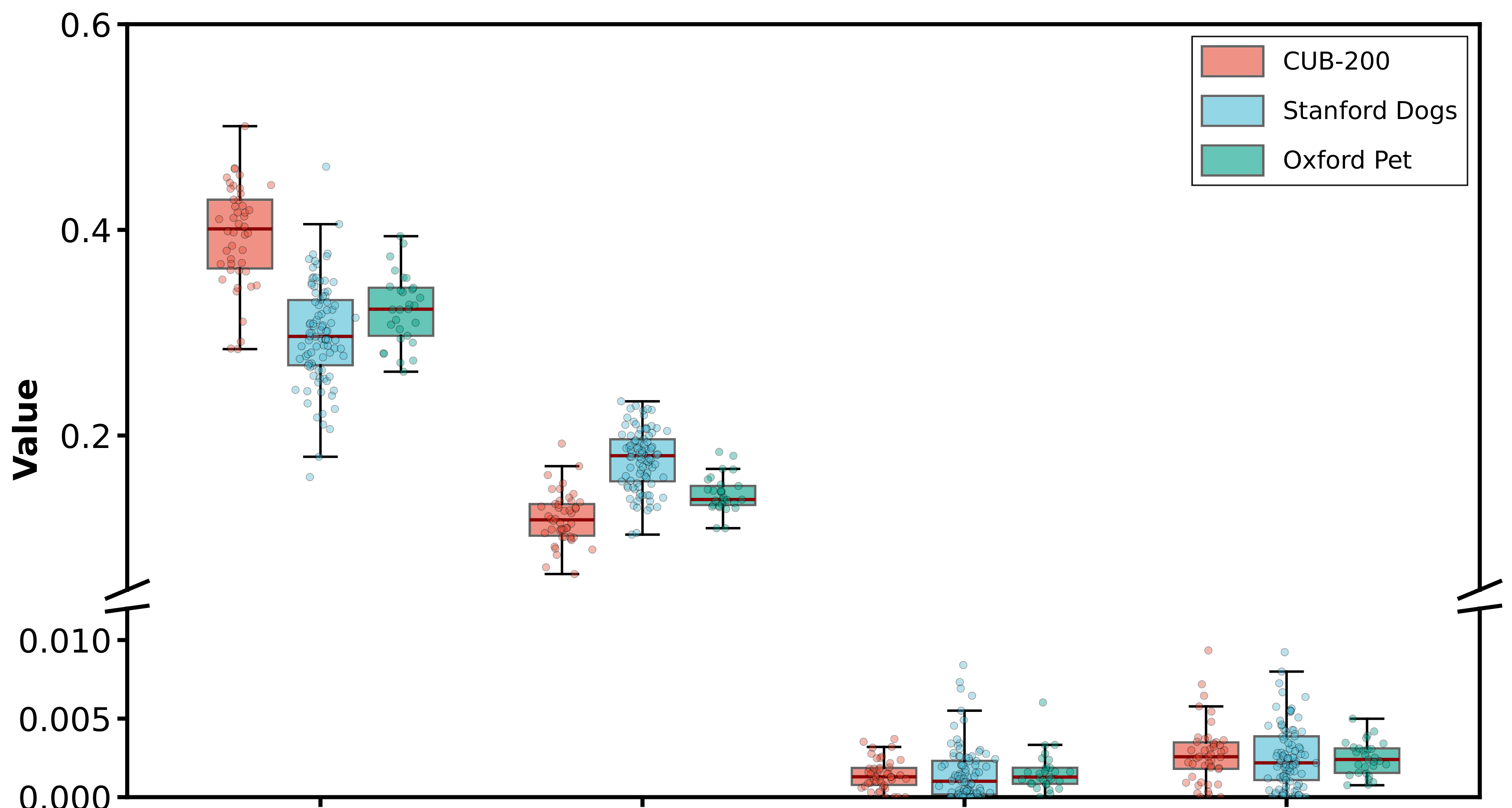}
{\scriptsize
\begin{tabular*}{0.98\columnwidth}{@{\extracolsep{\fill}}cccc}
$D(X_{\mathrm{fg}},R_{\mathrm{fg}})$ &
$D(X_{\mathrm{bg}},R_{\mathrm{bg}})$ &
$D(X_{\mathrm{fg}},R_{\mathrm{bg}})$ &
$D(X_{\mathrm{bg}},R_{\mathrm{fg}})$
\end{tabular*}}
\caption{Centered linear-kernel HSIC on three fine-grained datasets. Cross-region terms remain below $0.01$; $D(\cdot,\cdot)$ denotes dependence, not mutual information.}
\label{fig:dependence}
\end{figure}

\subsection{Classification Accuracy}
The reported S-IB point estimate exceeds its matched baseline in all 20 dataset--architecture pairs by 0.30--2.52 top-1 points (mean 0.91). The largest differences occur on CUB-200 and Stanford Dogs with ResNet-18/50 (Table~\ref{tab:classification}). Top-5 is higher in 15/20 pairs and lower in 5/20 (supplement).

\begin{table*}[t]
\centering
{\small
\begin{tabular}{lcccc}
\toprule
Dataset & ResNet-18 & ResNet-50 & DenseNet-121 & ViT-B/32 \\
\midrule
CIFAR-100 & 79.04$\rightarrow$79.59 & 80.16$\rightarrow$81.47 & 80.51$\rightarrow$80.91 & 78.77$\rightarrow$79.33 \\
ImageNet & 69.41$\rightarrow$70.38 & 75.72$\rightarrow$76.08 & 74.29$\rightarrow$74.81 & 74.84$\rightarrow$75.66 \\
CUB-200 & 69.15$\rightarrow$71.06 & 75.63$\rightarrow$78.15 & 74.16$\rightarrow$74.63 & 69.34$\rightarrow$70.16 \\
Oxford-IIIT Pet & 88.49$\rightarrow$88.89 & 91.93$\rightarrow$92.23 & 90.13$\rightarrow$90.98 & 88.93$\rightarrow$89.53 \\
Stanford Dogs & 74.59$\rightarrow$76.15 & 81.15$\rightarrow$82.86 & 78.40$\rightarrow$79.14 & 78.52$\rightarrow$79.41 \\
\bottomrule
\end{tabular}
}
\caption{Top-1 classification accuracy (\%). Entries show Baseline$\rightarrow$S-IB under matched training conditions.}
\label{tab:classification}
\end{table*}

\subsection{Localization across Explanation Methods}
Table~\ref{tab:localization_delta} averages the ResNet-50 localization change, S-IB minus baseline, over the three fine-grained datasets. Every method--metric mean is positive, but CAM-based methods gain about one mIoU point whereas gradient-only methods change little. This indicates compatibility, not uniform gains.

\begin{table}[t]
\centering
{\small
\begin{tabular}{lrrr}
\toprule
Method & $\Delta$Pixel Acc. & $\Delta$mIoU & $\Delta$mAP \\
\midrule
Saliency & +0.06 & +0.02 & +0.68 \\
Guided Backprop & +0.37 & +0.03 & +0.55 \\
Integrated Gradients & +0.07 & +0.03 & +1.02 \\
GradCAM & +1.02 & +1.04 & +1.82 \\
GradCAM++ & +0.88 & +0.88 & +0.82 \\
ScoreCAM & +0.89 & +0.89 & +1.47 \\
\bottomrule
\end{tabular}
}
\caption{Mean ResNet-50 localization change over three fine-grained datasets; full values are in the supplement.}
\label{tab:localization_delta}
\end{table}

ScoreCAM extends the positive observation beyond test-time-gradient explainers, but the experiment does not identify why CAM methods change more.

\begin{figure*}[t]
\centering
{\small
\begin{tabular*}{0.76\textwidth}{@{\extracolsep{\fill}}cccccccc}
Input & Saliency & Guided BP & IG & GradCAM & GradCAM++ & ScoreCAM & VJP
\end{tabular*}}
\includegraphics[width=0.76\textwidth]{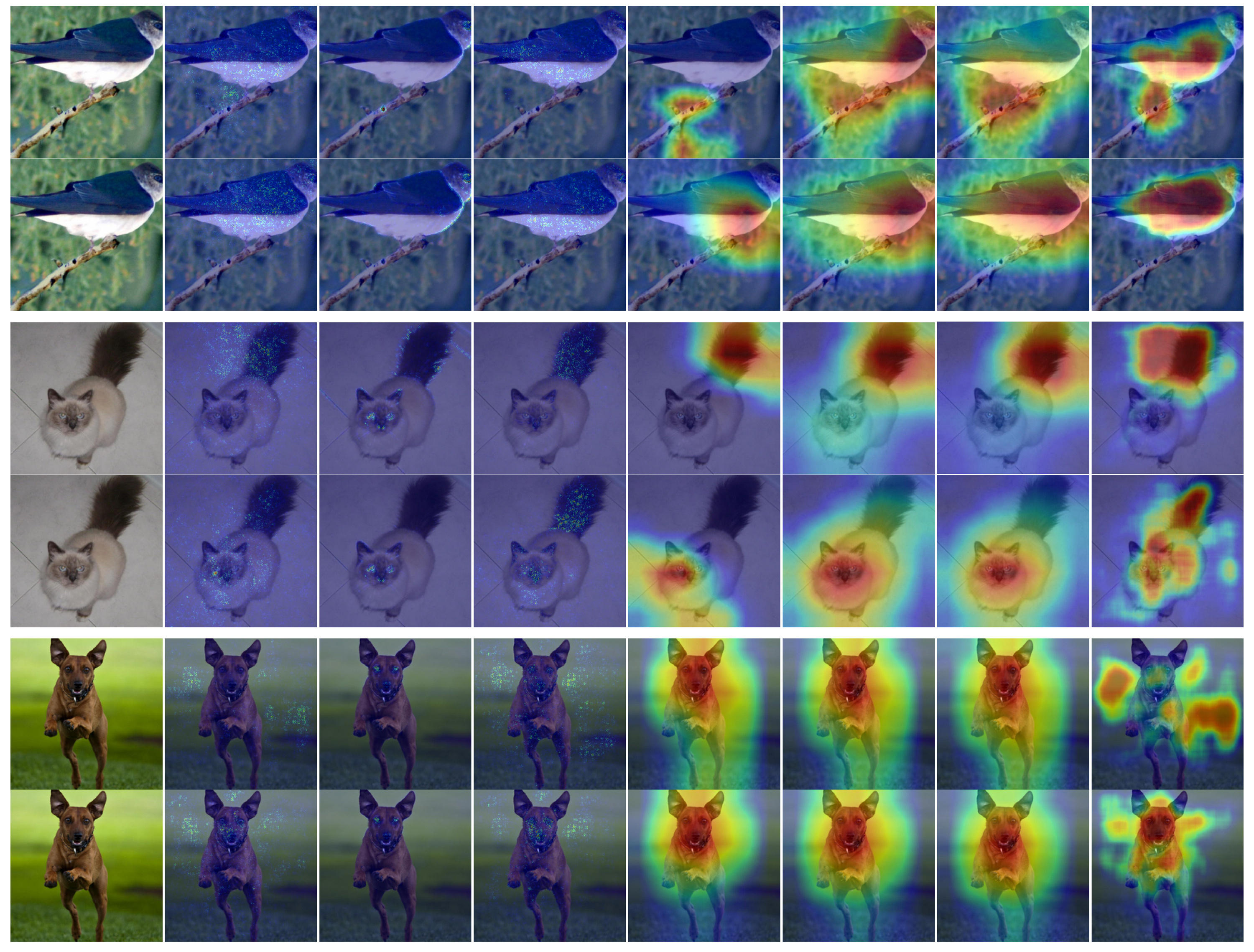}
\caption{Attribution maps on three datasets. Each pair shows the baseline (upper) and S-IB (lower). Responses are more object-centered after S-IB in these examples; tables provide quantitative evidence.}
\label{fig:qualitative}
\end{figure*}

\subsection{Faithfulness and Component Analysis}
All 12 architecture--dataset mean Insertion changes are positive (Table~\ref{tab:insertion}), ranging from +1.0 to +14.4 points. Mean Deletion is lower in 3/12 pairs and higher in 9/12; full per-signal values are in the supplement.

\begin{table}[t]
\centering
{\small
\begin{tabular}{lrrr}
\toprule
Backbone & CUB-200 & \shortstack{Oxford-IIIT\\Pet} & \shortstack{Stanford\\Dogs} \\
\midrule
ResNet-18 & +1.8 & +5.4 & +3.0 \\
ResNet-50 & +10.8 & +4.9 & +1.0 \\
DenseNet-121 & +14.4 & +5.7 & +14.2 \\
ViT-B/32 & +4.7 & +3.4 & +5.2 \\
\bottomrule
\end{tabular}
}
\caption{Seven-signal mean Insertion change.}
\label{tab:insertion}
\end{table}

Matched component controls probe the contributions of both objectives and VJP targeting (Table~\ref{tab:ablation}). In this setting, each objective exceeds the baseline, the full model is highest, and a random mask retains only part of the difference.

\begin{table}[t]
\centering
{\small
\begin{tabular}{lrr}
\toprule
Variant & Mean Insertion & Change \\
\midrule
Baseline & 48.5 & -- \\
Background only & 53.4 & +4.9 \\
Foreground only & 55.8 & +7.3 \\
Random mask & 51.7 & +3.2 \\
S-IB (full) & \textbf{59.2} & \textbf{+10.7} \\
\bottomrule
\end{tabular}
}
\caption{CUB-200/ResNet-50 component controls.}
\label{tab:ablation}
\end{table}

\subsection{Faithfulness Beyond Fine-Grained Data}
On CIFAR-100 and ImageNet with ResNet-50, seven-signal mean Insertion is higher, while Deletion, Infidelity, and Sensitivity are lower (Table~\ref{tab:general_faith}). These settings extend the favorable directional pattern beyond fine-grained data.

\begin{table}[t]
\centering
{\small
\setlength{\tabcolsep}{1mm}
\begin{tabular}{@{}lrrrr@{}}
\toprule
Dataset & Ins. $\uparrow$ & Del. $\downarrow$ & Infid. $\downarrow$ & Sens. $\downarrow$ \\
\midrule
CIFAR-100 & 27.8$\to$33.7 & 8.10$\to$7.97 & 260.1$\to$250.0 & 59.7$\to$53.9 \\
ImageNet & 48.6$\to$49.7 & 16.1$\to$15.4 & 199.6$\to$191.1 & 29.7$\to$28.7 \\
\bottomrule
\end{tabular}
}
\caption{Mean faithfulness over seven signals on ResNet-50. Infidelity is unscaled and descriptive because attribution magnitudes differ; other metrics are scaled by 100.}
\label{tab:general_faith}
\end{table}

\subsection{Robustness to Design Choices}
On CUB-200/ResNet-50, tested RBF bandwidths change mean Insertion by at most 0.6 points and Deletion by at most 0.2. Mask variants change Insertion by at most 0.9. Every setting remains above the 48.5 no-S-IB baseline within these ranges. The supplement reports all settings; only Insertion is available for mask variants.

\subsection{Foreground Coverage and Metric Profile}
Coverage, the mean retained-pixel count divided by $224^2$, differs markedly across datasets (Table~\ref{tab:coverage}). With ResNet-50, Insertion is higher for all seven signals on each dataset. Deletion improves for 7/7 on Stanford Dogs, 3/7 on CUB-200, and 0/7 on the near-full-support Oxford-IIIT Pet.

\begin{table}[t]
\centering
{\small
\setlength{\tabcolsep}{1mm}
\begin{tabular}{@{}lrrr@{}}
\toprule
Dataset & Coverage & Mean $\Delta$Ins. & Del. gain \\
\midrule
CUB-200 & 16.3\% & +10.8 & 3/7 \\
Oxford-IIIT Pet & 87.3\% & +4.9 & 0/7 \\
Stanford Dogs & 56.4\% & +1.0 & 7/7 \\
\bottomrule
\end{tabular}
}
\caption{Retained-pixel coverage and ResNet-50 faithfulness over seven attribution signals.}
\label{tab:coverage}
\end{table}

On Oxford-IIIT Pet, Insertion and Sensitivity improve for 7/7 signals and Infidelity for 6/7, but Deletion for 0/7; Section~\ref{sec:limitations} retains this adverse profile.

\subsection{Training Cost}
The VJP requires second-order differentiation during optimization. On one A100-80GB with ResNet-50, CUB-200 training increases from 43 to 105 seconds per epoch and peak memory from 4.5 to 9.2GB at batch size 64. ImageNet increases from 27 to 64 minutes per epoch and from 18.4 to 37.0GB at batch size 256. These measurements correspond to approximately $2.4\times$ time and $2.0\times$ memory. The unchanged backbone adds no inference operations or parameters.
 \section{Discussion and Limitations}
\label{sec:limitations}

\paragraph{What the evidence supports.}
Top-1 is higher in 20/20 matched settings, and mean Insertion is higher in all 14 reported settings. Top-5 is not uniformly higher. Mean Deletion is lower in 3/12 and higher in 9/12 fine-grained pairs, including 7/7 regressions on Oxford-IIIT Pet with ResNet-50. ScoreCAM and single-setting controls are consistent with training-time representation changes and contributions from both regularizers. They do not quantify retraining variance or identify a unique mechanism.

\paragraph{Posterior correctness and local scope.}
Exact recovery requires one piecewise-affine region, both geometric conditions, and $p_\theta=\eta$; standard calibration is weaker. Corollary~1 uses unmeasured posterior KL, so the experiments neither establish Bayes-posterior preservation nor verify all model-posterior recovery conditions. Ill-conditioning, saturation, activation boundaries, input-dependent masks, and smooth or training-coupled networks remain outside the exact claim. The rank study tests only injectivity.

\paragraph{Scope of the surrogates.}
Foreground HSIC measures dependence, not mutual information, and can reflect the shared gate. Random-mask controls favor VJP targeting but do not isolate gating. Cross-region HSIC supports approximate separation, while background variance leaves the all-ones component unconstrained. These diagnostics do not globalize the local theorem; S-IB is an optimization bias, not classical IB.

\paragraph{Applicability and cost.}
Evidence is limited to class-supervised image classification, and repeated-run variance is unavailable for some settings. Masks may fail for image-filling objects or ambiguous boundaries. Training costs $2.4\times$ time and $2.0\times$ memory; inference is unchanged.
\par
 \section{Conclusion}

S-IB uses posterior-concentration VJPs to reshape spatial evidence without changing the inference path. Under the stated local conditions, a fixed restricted VJP is posterior-equivalent and becomes minimally sufficient under posterior correctness; the learned mask is not claimed to recover a minimal support. The method keeps foreground and background pairs jointly reconstructible while applying asymmetric objectives. Conceptually, the analysis separates what a fixed support can retain from what the input-dependent mask is trained to encourage. The theorem supplies a local posterior-retention criterion, whereas HSIC and variance remain optimization surrogates rather than guarantees of global minimality or independence. Across matched evaluations, S-IB raises the reported Top-1 point estimates, ResNet-50 mean localization metrics, and mean Insertion for unchanged explainers using only class labels. Nonuniform Top-5 and Deletion results, unverified Bayes correctness, and second-order training cost bound the evidence. Thus, training-time evidence shaping complements unchanged post-hoc explanation rules.
\par

\begin{small}
\bibliography{references}
\end{small}

\section*{Supplementary Material}
\addcontentsline{toc}{section}{Supplementary Material}
\setcounter{section}{0}
\setcounter{subsection}{0}
\setcounter{subsubsection}{0}
\setcounter{figure}{0}
\setcounter{table}{0}
\setcounter{equation}{0}
\renewcommand{\thesection}{S\arabic{section}}
\renewcommand{\thefigure}{S\arabic{figure}}
\renewcommand{\thetable}{S\arabic{table}}
\renewcommand{\theequation}{S\arabic{equation}}
\raggedbottom

\makeatletter
\setlength{\@fptop}{0pt}
\setlength{\@fpsep}{12pt}
\setlength{\@fpbot}{0pt plus 1fil}
\makeatother
\renewcommand{\topfraction}{0.95}
\renewcommand{\bottomfraction}{0.95}
\renewcommand{\textfraction}{0.05}
\renewcommand{\floatpagefraction}{0.75}
\setlength{\textfloatsep}{12pt plus 2pt minus 2pt}
\setlength{\floatsep}{12pt plus 2pt minus 2pt}
\newenvironment{suppblock}%
  {\par\smallskip\noindent\begin{minipage}{\columnwidth}\centering}%
  {\end{minipage}\par\smallskip}\section{Implementation Details}

\subsection{VJP and Mask Construction}
For each input $x$, the implementation computes
\begin{equation}
R(x)=\nabla_x\frac{1}{2}\lVert p_\theta(x)\rVert_2^2=J_p(x)^\top p_\theta(x).
\end{equation}
This is a single VJP with the same shape as the input, not the unaggregated class-wise Jacobian tensor. Let $Q_q$ be the temperature-smoothed $q$-quantile and $\sigma_a(u)=\operatorname{sigmoid}(au)$. The implementation computes
\begin{equation}
\begin{aligned}
S&=\left|c^{-1}\textstyle\sum_jR_j\right|,\qquad
\widehat S=\frac{S-Q_{0.05}(S)}{Q_{0.95}(S)-Q_{0.05}(S)+\epsilon},\\
P&=\sigma_{10}(\widehat S-Q_{0.8}(\widehat S)),\qquad
D=\operatorname{AvgPool}_{31\times31}(P),\\
M&=\sigma_{10}(D-Q_{0.8}(D)).
\end{aligned}
\end{equation}
The quantile temperature is 0.1. The operator sorts each sample and applies a temperature-controlled weighted selection around the target index. Sorting is piecewise differentiable away from ties. The resulting spatial mask is broadcast across channels.

\subsection{Dependence Objective}
For flattened and normalized foreground tensors, the implementation constructs linear Gram matrices $K$ and $L$ and uses
\begin{equation}
D_{\mathrm{fg}}=\frac{1}{(B-1)^2}\operatorname{tr}(KHLH),
\quad H=I-\frac{1}{B}\mathbf{1}\mathbf{1}^\top.
\end{equation}
This is a centered linear-kernel HSIC estimator, not the finite-sample estimator commonly called unbiased HSIC. The background term is the mean per-sample variance of $R\odot(1-M)$ over all vectorized channel--spatial coordinates. The implementation optimizes
\begin{equation}
\mathcal{L}=\mathcal{L}_{\mathrm{ce}}+1000\exp(V_{\mathrm{bg}})+0.01\exp(-D_{\mathrm{fg}}).
\end{equation}
These coefficients and mask settings match the accompanying implementation.

\subsection{Coordinate-Centered Background-Variance Bound}
Let $r=\operatorname{vec}(R_{\mathrm{bg}})\in\mathbb{R}^{d}$ and $C_d=I-d^{-1}\mathbf{1}\mathbf{1}^{\top}$. The implementation uses the Bessel-corrected variance across vectorized channel--spatial coordinates,
\begin{equation}
v_{\mathrm{bg}}(r)=\frac{1}{d-1}\lVert C_dr\rVert_2^2,
\end{equation}
and $V_{\mathrm{bg}}$ is its batch mean. Define $\widetilde R_{\mathrm{bg}}=C_dr$ and $\widetilde\Sigma=\operatorname{Cov}(\widetilde R_{\mathrm{bg}})$. Assuming a finite second moment, let $N\sim\mathcal{N}(0,\sigma^2I_d)$ be independent. The Markov chain $X_{\mathrm{bg}}\to\widetilde R_{\mathrm{bg}}\to\widetilde R_{\mathrm{bg}}+N$, followed by the Gaussian maximum-entropy bound, gives
\begin{equation}
\begin{aligned}
I(X_{\mathrm{bg}};\widetilde R_{\mathrm{bg}}+N)
&\leq \frac{1}{2}\log\det\left(I+\frac{\widetilde\Sigma}{\sigma^2}\right)\\
&\leq \frac{\operatorname{tr}(\widetilde\Sigma)}{2\sigma^2}\\
&\leq \frac{(d-1)\mathbb{E}[v_{\mathrm{bg}}(r)]}{2\sigma^2}.
\end{aligned}
\end{equation}
The final inequality uses $\operatorname{tr}(\widetilde\Sigma)\leq\mathbb{E}\lVert\widetilde R_{\mathrm{bg}}\rVert_2^2=(d-1)\mathbb{E}[v_{\mathrm{bg}}(r)]$, because subtracting a mean cannot increase the second moment. This result applies to the coordinate-centered background VJP. Only the global all-ones component of the vectorized tensor lies in the null space of $C_d$ and is therefore unconstrained by this penalty.

\section{Proof Details for Local VJP Sufficiency}

\subsection{Factorization and Differential}
Inside a fixed activation region of a piecewise-affine classifier, write $z(x)=Ax+b$ and $p=\operatorname{softmax}(z/\tau)$. The softmax differential is
\begin{equation}
J_p(x)=\frac{1}{\tau}\left(\operatorname{diag}(p)-pp^\top\right)A.
\end{equation}
For a coordinate set $\mathcal{M}$, the restricted posterior-concentration VJP is therefore
\begin{equation}
R_{\mathcal{M}}=A_{\mathcal{M}}^\top g(p),\qquad
g(p)=\frac{1}{\tau}\left(p\odot p-\lVert p\rVert_2^2p\right).
\label{eq:supp_factor}
\end{equation}
For any simplex-tangent perturbation $v$ satisfying $\mathbf{1}^\top v=0$,
\begin{equation}
Dg_p[v]=\frac{1}{\tau}\left(2\operatorname{diag}(p)-2pp^\top-\lVert p\rVert_2^2I\right)v
=\frac{1}{\tau}B_pv.
\label{eq:supp_diff}
\end{equation}
Both $g(p)$ and $B_pv$ remain in the tangent space. At the uniform posterior, $B_pv=C^{-1}v$ on that space. Consequently, the determinant of $B_p|_{\mathcal{T}}$ is not identically zero as a function of $p$. Singular posteriors can therefore occupy only a lower-dimensional algebraic set. This genericity does not guarantee good numerical conditioning for a trained prediction.

\subsection{Local Invertibility and Minimality}
Let $F(p)=A_{\mathcal{M}}^\top g(p)$. At $p_0$, the differential restricted to the $(C-1)$-dimensional tangent space is
\begin{equation}
DF_{p_0}|_{\mathcal{T}}=\frac{1}{\tau}A_{\mathcal{M}}^\top B_{p_0}|_{\mathcal{T}}.
\end{equation}
If $B_{p_0}|_{\mathcal{T}}$ is nonsingular and $A_{\mathcal{M}}^\top$ is injective on $\mathcal{T}$, this differential has rank $C-1$. Continuity preserves that rank in a neighborhood of $p_0$. The constant-rank theorem then supplies local coordinates in which $F$ is the canonical embedding; equivalently, a smooth left inverse $h$ exists on the local image and satisfies $h(R_{\mathcal{M}})=p$.

Equation~\ref{eq:supp_factor} gives the reverse deterministic map $p\mapsto R_{\mathcal{M}}$. Hence $p$ and $R_{\mathcal{M}}$ generate the same local information. Let $P_{\mathcal V}$ denote the law conditioned on $X\in\mathcal V$. Under posterior correctness, $p(x)=\eta(x)=P_{\mathcal V}(Y=\cdot\mid X=x)$ for $x\in\mathcal V$. If a statistic $T(X)$ is sufficient under this conditional law, then
\begin{equation}
\eta(X)=P_{\mathcal V}(Y=\cdot\mid T(X))
\end{equation}
almost surely. The local inverse shows that $R_{\mathcal M}$ determines $\eta$, so the restricted VJP is sufficient. Conversely, every sufficient $T$ determines $\eta$ and therefore $R_{\mathcal M}=A_{\mathcal M}^{\top}g(\eta)$. Thus, $R_{\mathcal M}$ is a function of every sufficient statistic and is minimal sufficient under $P_{\mathcal V}$.

For spatial minimality, let $\Omega$ be the finite image grid and collect the supports satisfying the two geometric conditions at $(x_0,\mathcal U)$. Any nonempty such family has an inclusion-minimal element $\mathcal M^\star(x_0,\mathcal U)$. No proper subset satisfies the same local recoverability conditions, but an input in another activation region may admit a different minimal support. This combinatorial statement neither asserts a global support nor claims that the learned soft mask equals $\mathcal M^\star(x_0,\mathcal U)$.

\subsection{Approximate-Sufficiency Bound}
Let $\mathcal V$ be a neighborhood on which the left inverse above is valid. All information quantities and expectations in this subsection are under $P_{\mathcal V}$. Local equivalence gives $I(R_{\mathcal{M}};Y)=I(p;Y)$. Let $q(p)=P_{\mathcal V}(Y=\cdot\mid p)$. The information loss from replacing $X$ by $p$ is
\begin{align}
I(X;Y)-I(p;Y)
&=H(Y\mid p)-H(Y\mid X)\\
&=\mathbb{E}\left[D_{\mathrm{KL}}\left(\eta(X)\Vert q(p(X))\right)\right].
\end{align}
The conditional posterior $q$ minimizes expected log loss among all predictors measurable in $p$. Since $p$ itself is one such predictor,
\begin{equation}
I(X;Y)-I(R_{\mathcal{M}};Y)
\leq \mathbb{E}\left[D_{\mathrm{KL}}\left(\eta(X)\Vert p(X)\right)\right],
\end{equation}
which is Corollary~1. The bound can be loose, but it separates geometric non-invertibility from posterior approximation error.

\subsection{Boundaries of the Information-Theoretic Interpretation}
The exact theorem fixes one coordinate set throughout a neighborhood of a frozen piecewise-affine prediction map. The learned soft mask varies with the input and is therefore not itself covered by the inverse statement. Smooth GELU networks and training-time batch-coupled normalization receive only a first-order local interpretation. The foreground/background pairs reconstruct the original tensors exactly, but cross-region separation remains empirical. Both foreground variables share the VJP-derived gate, so their measured dependence can partly reflect common masking. The random-mask ablation does not isolate this contribution. The background bound concerns the Gaussian-smoothed, coordinate-centered quantity $I(X_{\mathrm{bg}};\widetilde R_{\mathrm{bg}}+N)$, not unsmoothed mutual information or the global all-ones component. Foreground HSIC measures statistical dependence and is not numerically interchangeable with mutual information.

\section{Complete Experimental Results and Diagnostics}

This section reports all recorded quantitative and qualitative results omitted from the main paper for space. All increases and decreases are retained, and Baseline$\to$S-IB denotes matched models before and after S-IB training. Earlier result tables used ``S-IB'' for the direct attribution signal; we report it consistently as ``Direct VJP'' to distinguish the signal from the training framework. All values are matched point estimates unless stated otherwise; no run-level significance claim is made.

\subsection{Full Localization Results on ResNet-50}
Table~\ref{tab:all_loc} reports all localization values used to construct the mean changes in the main paper. Entries are Baseline$\rightarrow$S-IB.

\begin{table*}[p]
\centering
{\small
\begin{tabular}{llccc}
\toprule
Dataset & Method & Pixel Acc. & mIoU & mAP \\
\midrule
CUB-200 & Saliency & 77.39$\to$77.43 & 38.87$\to$38.91 & 49.51$\to$50.14 \\
& Guided Backprop & 77.38$\to$78.41 & 38.82$\to$38.83 & 58.56$\to$59.25 \\
& Integrated Gradients & 78.21$\to$78.23 & 40.97$\to$40.98 & 56.49$\to$57.74 \\
& GradCAM & 77.10$\to$77.42 & 53.80$\to$54.18 & 52.84$\to$56.48 \\
& GradCAM++ & 75.43$\to$75.97 & 55.00$\to$55.43 & 62.06$\to$62.79 \\
& ScoreCAM & 75.31$\to$75.78 & 54.95$\to$55.36 & 63.40$\to$66.03 \\
& Direct VJP & 80.49$\to$80.82 & 58.41$\to$58.83 & 61.78$\to$63.92 \\
\midrule
Oxford-IIIT Pet & Saliency & 59.08$\to$59.13 & 29.66$\to$29.67 & 67.27$\to$68.46 \\
& Guided Backprop & 59.05$\to$59.09 & 29.60$\to$29.65 & 61.31$\to$62.21 \\
& Integrated Gradients & 59.06$\to$59.11 & 29.62$\to$29.70 & 64.86$\to$66.29 \\
& GradCAM & 70.72$\to$72.19 & 52.09$\to$53.67 & 71.42$\to$73.21 \\
& GradCAM++ & 71.71$\to$73.23 & 53.34$\to$54.96 & 73.44$\to$75.01 \\
& ScoreCAM & 73.01$\to$74.17 & 54.72$\to$56.01 & 75.14$\to$76.50 \\
& Direct VJP & 70.41$\to$71.07 & 51.86$\to$52.69 & 78.34$\to$79.53 \\
\midrule
Stanford Dogs & Saliency & 27.73$\to$27.82 & 13.94$\to$13.95 & 85.92$\to$86.13 \\
& Guided Backprop & 27.72$\to$27.77 & 13.86$\to$13.90 & 87.01$\to$87.06 \\
& Integrated Gradients & 27.67$\to$27.80 & 13.92$\to$13.93 & 86.32$\to$86.69 \\
& GradCAM & 59.19$\to$60.45 & 41.79$\to$42.95 & 91.21$\to$91.23 \\
& GradCAM++ & 60.67$\to$61.26 & 43.14$\to$43.73 & 91.43$\to$91.59 \\
& ScoreCAM & 60.01$\to$61.04 & 42.59$\to$43.55 & 91.25$\to$91.68 \\
& Direct VJP & 46.35$\to$46.81 & 30.56$\to$30.63 & 89.67$\to$89.91 \\
\bottomrule
\end{tabular}
}
\caption{Complete localization results on ResNet-50.}
\label{tab:all_loc}
\end{table*}

\subsection{Full Faithfulness Results on ResNet-50}
Table~\ref{tab:all_faith} shows Insertion and Deletion. Here and in every seven-signal average, the signals are the six post-hoc explainers plus Direct VJP. Oxford-IIIT Pet exhibits a consistent Deletion regression, which is retained rather than averaged away.

\begin{table*}[p]
\centering
{\small
\begin{tabular}{llcc}
\toprule
Dataset & Method & Insertion $\uparrow$ & Deletion $\downarrow$ \\
\midrule
CUB-200 & Saliency & 31.56$\to$36.92 & 5.63$\to$5.82 \\
& Guided Backprop & 42.12$\to$49.99 & 3.58$\to$4.60 \\
& Integrated Gradients & 42.80$\to$50.81 & 4.23$\to$5.63 \\
& GradCAM & 56.90$\to$70.07 & 12.79$\to$9.19 \\
& GradCAM++ & 56.68$\to$69.86 & 12.55$\to$8.70 \\
& ScoreCAM & 56.52$\to$70.02 & 12.02$\to$8.58 \\
& Direct VJP & 52.81$\to$67.08 & 10.56$\to$10.73 \\
\midrule
Oxford-IIIT Pet & Saliency & 36.05$\to$39.52 & 16.50$\to$19.48 \\
& Guided Backprop & 52.73$\to$54.38 & 10.61$\to$10.87 \\
& Integrated Gradients & 44.90$\to$47.84 & 13.32$\to$17.41 \\
& GradCAM & 69.02$\to$75.86 & 20.19$\to$22.49 \\
& GradCAM++ & 68.39$\to$75.59 & 19.94$\to$22.15 \\
& ScoreCAM & 68.06$\to$75.38 & 19.65$\to$21.55 \\
& Direct VJP & 58.45$\to$63.35 & 22.43$\to$24.56 \\
\midrule
Stanford Dogs & Saliency & 34.08$\to$34.31 & 6.30$\to$5.70 \\
& Guided Backprop & 44.81$\to$45.28 & 4.46$\to$3.97 \\
& Integrated Gradients & 41.58$\to$42.42 & 5.18$\to$4.38 \\
& GradCAM & 71.02$\to$72.27 & 11.35$\to$11.08 \\
& GradCAM++ & 70.60$\to$72.05 & 11.10$\to$10.95 \\
& ScoreCAM & 71.09$\to$72.78 & 11.01$\to$10.71 \\
& Direct VJP & 65.66$\to$66.47 & 13.78$\to$13.76 \\
\bottomrule
\end{tabular}
}
\caption{Complete faithfulness results on ResNet-50.}
\label{tab:all_faith}
\end{table*}

\begin{figure*}[t]
\begin{minipage}[t]{0.54\textwidth}
\subsection{Sensitivity Analysis}

The main paper reports the non-fine-grained faithfulness and training-cost values. On Oxford-IIIT Pet, all seven signals improve Insertion and Sensitivity, six improve Infidelity, and none improve Deletion. This profile motivates a bounded interpretation rather than a claim of uniform faithfulness improvement. Table~\ref{tab:robustness_supp} reports every recorded kernel and mask sensitivity setting. Its limited spread supports robustness only within these ranges; the mask sweep recorded Insertion but not Deletion.

\centering
{\small
\begin{tabular}{llrr}
\toprule
Family & Setting & Ins. $\uparrow$ & Del. $\downarrow$ \\
\midrule
No S-IB & -- & 48.5 & 8.8 \\
Kernel & Linear (default) & \textbf{59.2} & \textbf{7.6} \\
Kernel & RBF $0.5\times$ median & 58.6 & 7.8 \\
Kernel & RBF $1.0\times$ median & 59.1 & 7.7 \\
Kernel & RBF $2.0\times$ median & 58.8 & 7.7 \\
Mask slope & 5 / 20 & 58.4 / 58.9 & -- \\
Pool size & 21 / 41 & 58.3 / 58.7 & -- \\
Quantile & 0.7 / 0.9 & 58.5 / 58.6 & -- \\
\bottomrule
\end{tabular}
}
\captionof{table}{Kernel and mask sensitivity on CUB-200/ResNet-50. Values average seven attribution signals; mask variants report Insertion.}
\label{tab:robustness_supp}
\end{minipage}
\hfill
\begin{minipage}[t]{0.42\textwidth}
\subsection{Complete Cross-Architecture Results and Qualitative Diagnostics}

Table~\ref{tab:supp-classification-full} gives all Top-1 and Top-5 results, including small Top-5 decreases in saturated settings. Tables~\ref{tab:supp-resnet18}--\ref{tab:supp-vit} and~\ref{tab:supp-faithfulness-resnet18}--\ref{tab:supp-faithfulness-vit} provide every cross-architecture localization, Insertion, and Deletion entry. Figures~\ref{fig:supp-guided-original} and~\ref{fig:supp-dependence-original} provide additional qualitative diagnostics.
\end{minipage}
\end{figure*}

\begin{table*}[t]
\centering
\caption{Complete classification accuracy (\%). All recorded Top-1 and Top-5 values are reported. Columns show Baseline (B) and S-IB (S).}
\label{tab:supp-classification-full}
{\small
\setlength{\tabcolsep}{3pt}
\begin{tabular}{l|rrrr|rrrr}
\toprule
\multirow{2}{*}{Dataset} & \multicolumn{4}{c|}{ResNet-18} & \multicolumn{4}{c}{ResNet-50} \\
\cmidrule(lr){2-5}\cmidrule(l){6-9}
& B Top-1 & S Top-1 & B Top-5 & S Top-5
& B Top-1 & S Top-1 & B Top-5 & S Top-5 \\
\midrule
CIFAR-100 & 79.04 & 79.59 & 95.42 & 95.73 & 80.16 & 81.47 & 95.21 & 95.82 \\
ImageNet & 69.41 & 70.38 & 89.25 & 89.41 & 75.72 & 76.08 & 92.47 & 92.75 \\
CUB-200 & 69.15 & 71.06 & 90.61 & 91.09 & 75.63 & 78.15 & 92.85 & 94.58 \\
Oxford-IIIT Pet & 88.49 & 88.89 & 99.10 & 98.77 & 91.93 & 92.23 & 99.37 & 99.34 \\
Stanford Dogs & 74.59 & 76.15 & 95.75 & 96.17 & 81.15 & 82.86 & 97.94 & 98.01 \\
\bottomrule
\end{tabular}}

\vspace{0.8em}

{\small
\setlength{\tabcolsep}{3pt}
\begin{tabular}{l|rrrr|rrrr}
\toprule
\multirow{2}{*}{Dataset} & \multicolumn{4}{c|}{DenseNet-121} & \multicolumn{4}{c}{ViT-B/32} \\
\cmidrule(lr){2-5}\cmidrule(l){6-9}
& B Top-1 & S Top-1 & B Top-5 & S Top-5
& B Top-1 & S Top-1 & B Top-5 & S Top-5 \\
\midrule
CIFAR-100 & 80.51 & 80.91 & 95.96 & 96.18 & 78.77 & 79.33 & 94.72 & 95.75 \\
ImageNet & 74.29 & 74.81 & 91.74 & 92.29 & 74.84 & 75.66 & 91.26 & 92.31 \\
CUB-200 & 74.16 & 74.63 & 92.61 & 92.92 & 69.34 & 70.16 & 90.68 & 90.85 \\
Oxford-IIIT Pet & 90.13 & 90.98 & 99.26 & 99.32 & 88.93 & 89.53 & 99.01 & 98.96 \\
Stanford Dogs & 78.40 & 79.14 & 97.07 & 96.11 & 78.52 & 79.41 & 96.43 & 96.36 \\
\bottomrule
\end{tabular}}
\end{table*}

\begin{table*}[t]
\centering
\caption{Complete localization results on ResNet-18. Format: Baseline$\to$S-IB.}
\label{tab:supp-resnet18}
{\small
\setlength{\tabcolsep}{3pt}
\begin{tabular}{@{}lccc|ccc@{}}
\toprule
Method & \multicolumn{3}{c|}{CUB-200} & \multicolumn{3}{c}{Oxford-IIIT Pet} \\
\cmidrule(lr){2-4}\cmidrule(l){5-7}
& Pixel Acc. & mIoU & mAP & Pixel Acc. & mIoU & mAP \\
\midrule
Saliency & 78.35$\to$78.46 & 39.38$\to$39.40 & 40.69$\to$42.42 & 59.12$\to$59.13 & 29.72$\to$29.78 & 69.56$\to$69.57 \\
Guided Backprop & 78.34$\to$78.43 & 39.26$\to$39.29 & 58.03$\to$58.89 & 59.04$\to$59.07 & 29.58$\to$29.59 & 59.66$\to$59.98 \\
Integrated Gradients & 78.37$\to$78.41 & 39.37$\to$39.40 & 46.97$\to$48.61 & 59.09$\to$59.12 & 29.67$\to$29.69 & 65.54$\to$65.58 \\
GradCAM & 79.09$\to$79.55 & 58.24$\to$58.79 & 62.83$\to$63.90 & 73.94$\to$74.23 & 55.60$\to$56.10 & 76.76$\to$76.87 \\
GradCAM++ & 77.30$\to$77.98 & 57.81$\to$58.55 & 69.54$\to$70.34 & 74.98$\to$75.26 & 57.10$\to$57.32 & 78.10$\to$79.24 \\
ScoreCAM & 78.26$\to$78.93 & 58.74$\to$59.48 & 70.07$\to$70.83 & 75.23$\to$75.72 & 57.28$\to$57.69 & 78.60$\to$79.45 \\
Direct VJP & 78.87$\to$79.80 & 55.36$\to$56.73 & 53.89$\to$56.27 & 73.09$\to$73.21 & 54.99$\to$55.38 & 83.42$\to$83.48 \\
\bottomrule
\end{tabular}

\vspace{0.6em}

\begin{tabular}{@{}lccc@{}}
\toprule
Method & \multicolumn{3}{c}{Stanford Dogs} \\
\cmidrule(l){2-4}
& Pixel Acc. & mIoU & mAP \\
\midrule
Saliency & 27.09$\to$27.17 & 13.59$\to$13.71 & 87.83$\to$87.89 \\
Guided Backprop & 26.99$\to$27.21 & 13.51$\to$13.63 & 87.35$\to$87.68 \\
Integrated Gradients & 27.05$\to$27.32 & 13.56$\to$13.59 & 87.12$\to$87.27 \\
GradCAM & 57.66$\to$57.89 & 40.28$\to$40.43 & 91.69$\to$91.95 \\
GradCAM++ & 59.91$\to$60.03 & 42.34$\to$42.45 & 91.90$\to$92.34 \\
ScoreCAM & 61.67$\to$61.78 & 43.97$\to$44.07 & 92.07$\to$92.43 \\
Direct VJP & 46.51$\to$46.69 & 30.28$\to$30.36 & 90.31$\to$90.65 \\
\bottomrule
\end{tabular}}
\end{table*}

\begin{table*}[t]
\centering
\caption{Complete localization results on DenseNet-121. Format: Baseline$\to$S-IB.}
\label{tab:supp-densenet121}
{\small
\setlength{\tabcolsep}{3pt}
\begin{tabular}{@{}lccc|ccc@{}}
\toprule
Method & \multicolumn{3}{c|}{CUB-200} & \multicolumn{3}{c}{Oxford-IIIT Pet} \\
\cmidrule(lr){2-4}\cmidrule(l){5-7}
& Pixel Acc. & mIoU & mAP & Pixel Acc. & mIoU & mAP \\
\midrule
Saliency & 78.19$\to$78.20 & 39.33$\to$39.33 & 48.41$\to$49.32 & 59.11$\to$59.12 & 29.70$\to$29.72 & 68.95$\to$69.23 \\
Guided Backprop & 78.18$\to$78.20 & 39.20$\to$39.25 & 64.81$\to$65.19 & 59.04$\to$59.07 & 29.58$\to$29.63 & 64.18$\to$64.79 \\
Integrated Gradients & 78.22$\to$78.25 & 39.36$\to$39.41 & 61.94$\to$62.91 & 59.09$\to$59.15 & 29.67$\to$29.69 & 67.86$\to$68.45 \\
GradCAM & 81.36$\to$81.52 & 59.97$\to$60.24 & 70.68$\to$72.10 & 75.20$\to$75.31 & 56.97$\to$57.16 & 78.42$\to$78.97 \\
GradCAM++ & 80.76$\to$81.34 & 59.91$\to$60.73 & 72.26$\to$73.94 & 75.74$\to$75.88 & 57.74$\to$57.93 & 79.04$\to$79.67 \\
ScoreCAM & 79.54$\to$80.15 & 58.11$\to$59.82 & 68.60$\to$69.47 & 74.80$\to$75.12 & 56.70$\to$56.91 & 77.28$\to$77.83 \\
Direct VJP & 79.11$\to$79.38 & 56.81$\to$57.13 & 62.87$\to$63.82 & 71.37$\to$71.43 & 52.96$\to$53.26 & 80.05$\to$80.21 \\
\bottomrule
\end{tabular}

\vspace{0.6em}

\begin{tabular}{@{}lccc@{}}
\toprule
Method & \multicolumn{3}{c}{Stanford Dogs} \\
\cmidrule(l){2-4}
& Pixel Acc. & mIoU & mAP \\
\midrule
Saliency & 27.86$\to$27.85 & 13.98$\to$13.99 & 86.22$\to$87.13 \\
Guided Backprop & 27.76$\to$27.77 & 13.90$\to$13.90 & 89.15$\to$90.04 \\
Integrated Gradients & 27.82$\to$27.81 & 13.95$\to$13.96 & 87.01$\to$87.67 \\
GradCAM & 51.59$\to$51.70 & 35.42$\to$35.61 & 93.55$\to$93.95 \\
GradCAM++ & 52.74$\to$52.83 & 36.40$\to$36.63 & 93.60$\to$94.01 \\
ScoreCAM & 52.20$\to$52.45 & 35.79$\to$36.92 & 93.13$\to$93.78 \\
Direct VJP & 47.23$\to$47.59 & 32.06$\to$32.41 & 93.31$\to$93.62 \\
\bottomrule
\end{tabular}}
\end{table*}

\begin{table*}[t]
\centering
\caption{Complete localization results on ViT-B/32. Format: Baseline$\to$S-IB.}
\label{tab:supp-vit}
{\small
\setlength{\tabcolsep}{3pt}
\begin{tabular}{@{}lccc|ccc@{}}
\toprule
Method & \multicolumn{3}{c|}{CUB-200} & \multicolumn{3}{c}{Oxford-IIIT Pet} \\
\cmidrule(lr){2-4}\cmidrule(l){5-7}
& Pixel Acc. & mIoU & mAP & Pixel Acc. & mIoU & mAP \\
\midrule
Saliency & 78.31$\to$78.32 & 39.79$\to$39.80 & 57.85$\to$58.56 & 59.17$\to$59.18 & 29.84$\to$29.86 & 65.42$\to$65.97 \\
Guided Backprop & 78.34$\to$78.34 & 39.80$\to$39.92 & 57.83$\to$68.93 & 59.19$\to$59.23 & 29.87$\to$29.87 & 65.44$\to$65.98 \\
Integrated Gradients & 78.28$\to$78.28 & 39.61$\to$39.62 & 64.86$\to$65.24 & 59.12$\to$59.13 & 29.71$\to$29.75 & 61.05$\to$62.19 \\
GradCAM & 65.40$\to$65.87 & 36.32$\to$36.41 & 34.79$\to$35.35 & 51.66$\to$52.58 & 28.01$\to$28.69 & 44.57$\to$44.60 \\
GradCAM++ & 64.54$\to$65.79 & 35.85$\to$36.02 & 35.41$\to$35.92 & 51.08$\to$52.53 & 27.62$\to$28.40 & 43.90$\to$44.86 \\
ScoreCAM & 81.33$\to$81.42 & 54.35$\to$55.13 & 66.24$\to$66.83 & 64.16$\to$64.60 & 39.69$\to$39.97 & 67.14$\to$67.29 \\
Direct VJP & 83.95$\to$84.23 & 63.64$\to$64.29 & 74.87$\to$75.92 & 71.08$\to$71.35 & 52.06$\to$52.38 & 79.15$\to$79.72 \\
\bottomrule
\end{tabular}

\vspace{0.6em}

\begin{tabular}{@{}lccc@{}}
\toprule
Method & \multicolumn{3}{c}{Stanford Dogs} \\
\cmidrule(l){2-4}
& Pixel Acc. & mIoU & mAP \\
\midrule
Saliency & 27.92$\to$27.93 & 14.05$\to$14.06 & 85.77$\to$85.93 \\
Guided Backprop & 27.93$\to$27.95 & 14.06$\to$14.09 & 85.78$\to$85.95 \\
Integrated Gradients & 27.85$\to$27.87 & 13.98$\to$14.05 & 84.23$\to$85.04 \\
GradCAM & 40.59$\to$40.96 & 22.48$\to$22.52 & 73.06$\to$73.68 \\
GradCAM++ & 38.76$\to$39.07 & 21.11$\to$21.42 & 71.79$\to$71.93 \\
ScoreCAM & 33.40$\to$33.92 & 18.99$\to$18.70 & 85.56$\to$85.63 \\
Direct VJP & 46.36$\to$46.58 & 31.44$\to$31.67 & 93.02$\to$93.25 \\
\bottomrule
\end{tabular}}
\end{table*}

\begin{table*}[t]
\centering
\caption{Complete faithfulness results on ResNet-18. Format: Baseline$\to$S-IB.}
\label{tab:supp-faithfulness-resnet18}
{\small
\setlength{\tabcolsep}{3pt}
\begin{tabular}{l|cc|cc|cc}
\toprule
\multirow{2}{*}{Method} & \multicolumn{2}{c|}{CUB-200} & \multicolumn{2}{c|}{Oxford-IIIT Pet} & \multicolumn{2}{c}{Stanford Dogs} \\
\cmidrule{2-7}
& Insertion $\uparrow$ & Deletion $\downarrow$ & Insertion $\uparrow$ & Deletion $\downarrow$ & Insertion $\uparrow$ & Deletion $\downarrow$ \\
\midrule
Saliency & 12.42$\to$14.58 & 3.37$\to$3.46 & 24.15$\to$27.73 & 10.35$\to$10.90 & 20.60$\to$21.92 & 3.46$\to$3.73 \\
Guided Backprop & 17.33$\to$19.12 & 2.75$\to$2.72 & 33.74$\to$35.93 & 7.34$\to$7.98 & 26.55$\to$28.70 & 3.48$\to$3.77 \\
Integrated Gradients & 19.81$\to$21.81 & 2.53$\to$2.65 & 30.55$\to$34.96 & 9.02$\to$9.47 & 24.62$\to$27.21 & 3.35$\to$3.58 \\
GradCAM & 29.22$\to$30.61 & 4.21$\to$4.29 & 58.67$\to$65.79 & 18.73$\to$18.15 & 55.64$\to$59.12 & 7.83$\to$9.78 \\
GradCAM++ & 28.61$\to$29.98 & 4.12$\to$4.21 & 57.01$\to$63.78 & 16.57$\to$18.42 & 54.49$\to$58.24 & 7.89$\to$9.93 \\
ScoreCAM & 28.79$\to$30.13 & 4.10$\to$4.19 & 57.26$\to$64.16 & 16.87$\to$18.42 & 56.11$\to$59.85 & 7.40$\to$9.28 \\
Direct VJP & 25.98$\to$28.17 & 5.55$\to$5.64 & 47.90$\to$54.50 & 17.45$\to$19.32 & 51.31$\to$54.97 & 8.91$\to$10.63 \\
\bottomrule
\end{tabular}}
\end{table*}

\begin{table*}[t]
\centering
\caption{Complete faithfulness results on DenseNet-121. Format: Baseline$\to$S-IB.}
\label{tab:supp-faithfulness-densenet121}
{\small
\setlength{\tabcolsep}{3pt}
\begin{tabular}{l|cc|cc|cc}
\toprule
\multirow{2}{*}{Method} & \multicolumn{2}{c|}{CUB-200} & \multicolumn{2}{c|}{Oxford-IIIT Pet} & \multicolumn{2}{c}{Stanford Dogs} \\
\cmidrule{2-7}
& Insertion $\uparrow$ & Deletion $\downarrow$ & Insertion $\uparrow$ & Deletion $\downarrow$ & Insertion $\uparrow$ & Deletion $\downarrow$ \\
\midrule
Saliency & 16.83$\to$26.57 & 3.83$\to$4.37 & 28.42$\to$32.53 & 12.12$\to$15.01 & 32.99$\to$43.61 & 5.21$\to$11.60 \\
Guided Backprop & 26.62$\to$40.06 & 3.05$\to$3.20 & 44.40$\to$49.82 & 9.09$\to$9.83 & 43.49$\to$55.01 & 5.37$\to$12.46 \\
Integrated Gradients & 26.51$\to$39.60 & 3.24$\to$3.45 & 39.73$\to$44.93 & 12.48$\to$13.36 & 37.59$\to$52.01 & 5.25$\to$10.15 \\
GradCAM & 46.10$\to$63.36 & 5.95$\to$7.16 & 62.79$\to$69.65 & 16.41$\to$17.66 & 63.05$\to$79.75 & 8.54$\to$12.80 \\
GradCAM++ & 45.40$\to$62.02 & 6.20$\to$7.49 & 61.84$\to$68.65 & 16.61$\to$17.87 & 62.54$\to$78.87 & 8.73$\to$13.19 \\
ScoreCAM & 45.30$\to$61.99 & 6.72$\to$8.40 & 61.37$\to$67.68 & 17.95$\to$20.24 & 62.21$\to$76.37 & 9.93$\to$15.44 \\
Direct VJP & 39.74$\to$53.53 & 9.44$\to$11.28 & 53.82$\to$58.92 & 20.74$\to$22.45 & 58.49$\to$73.79 & 9.70$\to$17.13 \\
\bottomrule
\end{tabular}}
\end{table*}

\begin{table*}[t]
\centering
\caption{Complete faithfulness results on ViT-B/32. Format: Baseline$\to$S-IB.}
\label{tab:supp-faithfulness-vit}
{\small
\setlength{\tabcolsep}{3pt}
\begin{tabular}{l|cc|cc|cc}
\toprule
\multirow{2}{*}{Method} & \multicolumn{2}{c|}{CUB-200} & \multicolumn{2}{c|}{Oxford-IIIT Pet} & \multicolumn{2}{c}{Stanford Dogs} \\
\cmidrule{2-7}
& Insertion $\uparrow$ & Deletion $\downarrow$ & Insertion $\uparrow$ & Deletion $\downarrow$ & Insertion $\uparrow$ & Deletion $\downarrow$ \\
\midrule
Saliency & 34.68$\to$39.82 & 4.35$\to$4.80 & 50.19$\to$52.34 & 17.35$\to$16.07 & 37.68$\to$43.41 & 8.10$\to$9.11 \\
Guided Backprop & 34.73$\to$40.03 & 4.39$\to$4.97 & 50.23$\to$52.47 & 17.33$\to$16.35 & 37.73$\to$43.47 & 8.12$\to$9.15 \\
Integrated Gradients & 39.90$\to$45.52 & 3.28$\to$3.68 & 56.20$\to$57.98 & 13.16$\to$12.23 & 44.78$\to$50.52 & 5.48$\to$6.64 \\
GradCAM & 26.63$\to$30.07 & 15.88$\to$19.48 & 45.31$\to$47.32 & 40.36$\to$40.08 & 30.02$\to$33.50 & 26.08$\to$27.24 \\
GradCAM++ & 26.24$\to$29.85 & 16.30$\to$19.09 & 45.14$\to$58.14 & 39.80$\to$39.58 & 29.59$\to$33.17 & 25.73$\to$26.81 \\
ScoreCAM & 38.13$\to$42.90 & 5.55$\to$6.02 & 60.83$\to$62.47 & 23.74$\to$23.68 & 49.16$\to$52.26 & 14.93$\to$17.64 \\
Direct VJP & 36.51$\to$41.19 & 4.39$\to$4.71 & 58.84$\to$59.64 & 19.75$\to$20.09 & 46.74$\to$55.65 & 12.04$\to$11.65 \\
\bottomrule
\end{tabular}}
\end{table*}

\begin{figure*}[t]
\centering
\includegraphics[width=0.86\textwidth]{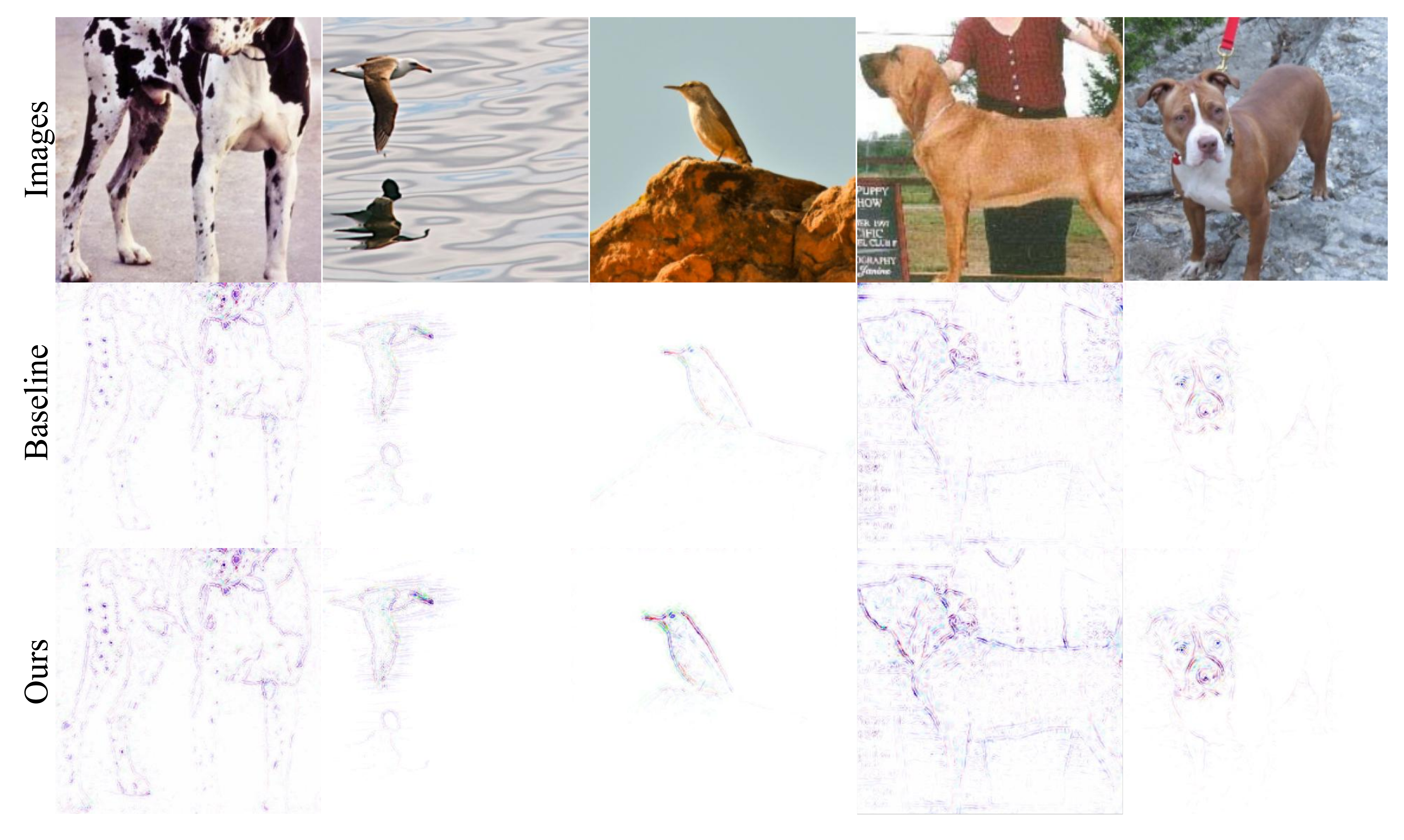}
\caption{Additional Guided Backpropagation diagnostic on ResNet-50. Rows show the input, baseline response, S-IB response, and their difference. Red and blue indicate positive and negative changes, respectively. These examples suggest sharper object boundaries and reduced background responses, but do not establish a population-level or causal effect.}
\label{fig:supp-guided-original}
\end{figure*}

\begin{figure*}[t]
\centering
\includegraphics[width=0.86\textwidth]{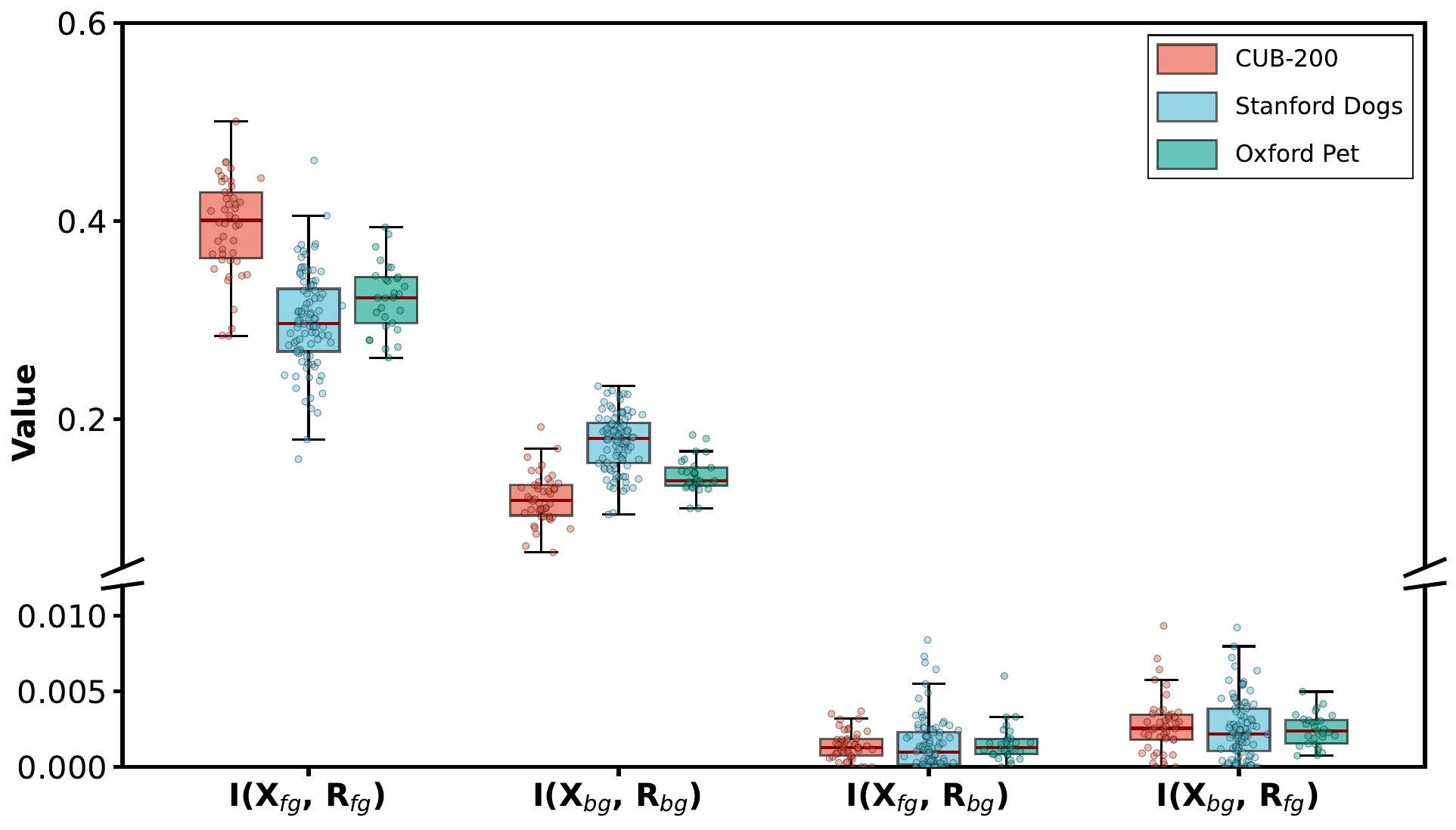}
\caption{Additional four-panel dependence plot. The embedded artwork uses mutual-information terminology, but the plotted quantities are centered linear-kernel HSIC estimates. The figure therefore supports a descriptive cross-versus-within dependence comparison, not a numerical mutual-information estimate or proof of independence.}
\label{fig:supp-dependence-original}
\end{figure*}

\FloatBarrier
 
\begin{figure*}[p]
\begin{minipage}[t]{0.48\textwidth}
\subsection{Normalized-Dependence and Cross-Architecture Summaries}

The additional analyses include a transformed dependence diagnostic and cross-architecture separation summaries. The normalized score is neither mutual information nor comparable across datasets. Because attribution methods are not independent training repeats, their dispersion does not support run-level significance inference.

\centering
{\small
\setlength{\tabcolsep}{3pt}
\begin{tabular}{lll}
\toprule
Backbone & Dataset & Baseline$\to$S-IB \\
\midrule
ResNet-18 & CUB-200 & 0.510$\to$0.614 \\
ResNet-18 & Oxford-IIIT Pet & 1.732$\to$1.914 \\
ResNet-18 & Stanford Dogs & 1.435$\to$1.455 \\
ResNet-50 & CUB-200 & 0.415$\to$1.191 \\
ResNet-50 & Oxford-IIIT Pet & 0.327$\to$0.393 \\
ResNet-50 & Stanford Dogs & 0.207$\to$0.237 \\
DenseNet-121 & CUB-200 & 0.392$\to$0.613 \\
DenseNet-121 & Oxford-IIIT Pet & 0.343$\to$0.483 \\
DenseNet-121 & Stanford Dogs & 0.768$\to$1.045 \\
ViT-B/32 & CUB-200 & 5.031$\to$5.307 \\
ViT-B/32 & Oxford-IIIT Pet & 3.728$\to$3.771 \\
ViT-B/32 & Stanford Dogs & 3.465$\to$3.494 \\
\bottomrule
\end{tabular}}
\captionof{table}{Normalized-dependence diagnostic, reported as Baseline$\to$S-IB for $\exp(\operatorname{zscore}(D_{\mathrm{fg}}-D_{\mathrm{bg}}))$ within each dataset. This descriptive transform is neither mutual information nor comparable across datasets.}
\label{tab:supp-legacy-info-diff}

\medskip

{\small
\setlength{\tabcolsep}{2.5pt}
\begin{tabular}{llrrr}
\toprule
Backbone & Dataset & $\Delta$Ins. & Relative & $d$ \\
\midrule
ResNet-18 & CUB-200 & +1.8 & +7.5\% & 4.60 \\
ResNet-18 & Oxford-IIIT Pet & +5.4 & +12.1\% & 2.73 \\
ResNet-18 & Stanford Dogs & +3.0 & +7.1\% & 3.09 \\
ResNet-50 & CUB-200 & +10.8 & +22.2\% & 3.01 \\
ResNet-50 & Oxford-IIIT Pet & +4.9 & +8.6\% & 2.14 \\
ResNet-50 & Stanford Dogs & +1.0 & +1.7\% & 1.83 \\
DenseNet-121 & CUB-200 & +14.4 & +40.8\% & 5.37 \\
DenseNet-121 & Oxford-IIIT Pet & +5.7 & +11.3\% & 5.61 \\
DenseNet-121 & Stanford Dogs & +14.2 & +27.5\% & 6.12 \\
ViT-B/32 & CUB-200 & +4.7 & +13.7\% & 5.58 \\
ViT-B/32 & Oxford-IIIT Pet & +3.4 & +6.4\% & 0.79 \\
ViT-B/32 & Stanford Dogs & +5.2 & +13.2\% & 2.56 \\
\bottomrule
\end{tabular}}
\captionof{table}{Cross-architecture Insertion summary averaged over seven attribution signals. Because the signals are not repeated training runs, Cohen's $d$ is a descriptive separation index, not a run-level effect size.}
\label{tab:supp-legacy-effect}
\end{minipage}
\hfill
\begin{minipage}[t]{0.48\textwidth}
\centering
{\small
\setlength{\tabcolsep}{2pt}
\begin{tabular}{lrrrr}
\toprule
Dataset & $\Delta$mIoU & $\Delta$Ins. & Ratio & $p$ \\
\midrule
CUB-200 & $+0.24\pm0.21$ & $+10.77\pm3.57$ & $44.9\times$ & $<10^{-4}$ \\
Oxford-IIIT Pet & $+0.78\pm0.73$ & $+4.90\pm2.29$ & $6.3\times$ & $<10^{-3}$ \\
Stanford Dogs & $+0.41\pm0.50$ & $+0.96\pm0.53$ & $2.3\times$ & $<10^{-3}$ \\
\bottomrule
\end{tabular}}
\captionof{table}{ResNet-50 asymmetry diagnostic over seven attribution methods. The methods are neither independent samples nor repeated runs; paired $p$-values are descriptive and are not used for inference.}
\label{tab:supp-legacy-asymmetry}

\subsection{Additional Recorded Analyses}

\raggedright
The additional experiments cover training cost, four-metric faithfulness, foreground coverage, sensitivity, component controls, restricted-Jacobian rank, and cross-/within-region HSIC. Values appear in the main paper or Table~\ref{tab:robustness_supp}; Table~\ref{tab:supp-rebuttal-counts} gives the per-metric improved-signal counts.

\centering
{\small
\setlength{\tabcolsep}{2pt}
\begin{tabular}{lrrrr}
\toprule
Dataset & Ins. & Del. & Infid. & Sens. \\
\midrule
CIFAR-100 & 7/7 & 4/7 & 7/7 & 6/7 \\
ImageNet & 7/7 & 6/7 & 7/7 & 6/7 \\
Oxford-IIIT Pet & 7/7 & 0/7 & 6/7 & 7/7 \\
\bottomrule
\end{tabular}}
\captionof{table}{Improved attribution signals (out of seven) in the additional analyses.}
\label{tab:supp-rebuttal-counts}

\raggedright
On CIFAR-100, Insertion increased by approximately 84--96\% for gradient-based signals and 12--13\% for CAM-based signals. Guided Backpropagation was the only recorded ImageNet Deletion exception ($7.60\to8.03$). On Oxford-IIIT Pet, representative Infidelity changed from $400.45$ to $333.64$ for GradCAM and from $436.34$ to $388.96$ for GradCAM++. Complete per-signal CIFAR-100/ImageNet Infidelity and Sensitivity tables were unavailable, so no missing values are reconstructed.

In the HSIC-dependence-versus-epoch plot, the transition near epoch 23 does not coincide with the recorded learning-rate reduction at epoch 30; foreground and background inflections differ by one to two epochs. A preliminary bootstrap check kept within-region estimates above ten times the cross-region estimates across dataset splits. Without interval endpoints, no reproducible confidence interval is available.
\end{minipage}
\end{figure*}
  
\end{document}